\RequirePackage{snapshot}
\RequirePackage{snapshot}
\documentclass{article}

\usepackage{times}
\usepackage{graphicx}
\usepackage{sidecap}
\usepackage[small]{caption}
\usepackage{subcaption}
\usepackage{amsmath}
\allowdisplaybreaks
\usepackage{amsthm}

\usepackage{thmtools}
\usepackage{thm-restate}

\usepackage{multicol}

\usepackage{pifont}
\usepackage{xspace}

\usepackage[round]{natbib}

\usepackage{appendix}

\newcommand{\rightcomment}[1]{\(\triangleright\) {\small \it #1}}
\newcommand{\eqcomment}[1]{\addtocounter{equation}{1}\tag*{\rightcomment{#1}\quad(\theequation)}}
\usepackage{suffix}
\WithSuffix\newcommand\eqcomment*[1]{\tag*{\rightcomment{#1}}}

\usepackage[hidelinks]{hyperref}

\usepackage[final]{neurips_2022}

\usepackage[disable]{todonotes}

\usepackage[utf8]{inputenc}
\usepackage[T1]{fontenc}
\usepackage{hyperref}
\usepackage{url}
\usepackage{booktabs}
\usepackage{amsfonts}
\usepackage{nicefrac}
\usepackage{microtype}
\usepackage{xcolor}

\usepackage{algorithm}
\usepackage[noend]{algpseudocode}

\renewcommand\algorithmicdo{:}
\renewcommand\algorithmicthen{:}
\algnewcommand{\IfThen}[2]{\State \algorithmicif\ #1\ \algorithmicthen\ #2}
\algnewcommand{\IfThenElse}[3]{\State \algorithmicif\ #1\ \algorithmicthen\ #2\ \algorithmicelse\ #3}
\algrenewcommand{\algorithmiccomment}[1]{\hfill \rightcomment{#1}}
\algnewcommand{\LineComment}[1]{\State \rightcomment{#1}}
\algnewcommand{\LinesComment}[1]{\State \rightcomment{\parbox[t]{\linewidth-\leftmargin-\widthof{\(\triangleright\) }}{#1}}\smallskip}
\algnewcommand\algorithmicinput{{\bfseries Input:}}
\algnewcommand\INPUT{\item[\algorithmicinput]}
\algnewcommand\algorithmicoutput{{\bfseries Output:}}
\algnewcommand\OUTPUT{\item[\algorithmicoutput]}
\makeatletter
\newcounter{algorithmicH}
\let\oldalgorithmic\algorithmic
\renewcommand{\algorithmic}{%
  \stepcounter{algorithmicH}
  \oldalgorithmic}
\renewcommand{\theHALG@line}{ALG@line.\thealgorithmicH.\arabic{ALG@line}}
\makeatother
\makeatletter
\newcommand{\algmargin}{\the\ALG@thistlm}
\makeatother
\algnewcommand{\Statepar}[1]{\State\parbox[t]{\dimexpr\linewidth-\algmargin}{\strut #1\strut}}

\newcommand{\pluseq}{\mathrel{+\!\!=}}

\usepackage{xpatch}

\usepackage[compact]{titlesec}
\titlespacing{\section}{0pt}{1ex}{0.5ex}
\titlespacing{\subsection}{0pt}{0.5ex}{0ex}
\titlespacing{\subsubsection}{0pt}{0.5ex}{0ex} 
\renewcommand{\paragraph}[1]{\textbf{#1}}
\usepackage{setspace}

\AtBeginDocument{%
  \addtolength\abovedisplayskip{-0.25\baselineskip}%
  \addtolength\belowdisplayskip{-0.25\baselineskip}%
  \addtolength\abovedisplayshortskip{-0.25\baselineskip}%
  \addtolength\belowdisplayshortskip{-0.25\baselineskip}%
}

\usepackage{amsmath}
\usepackage{amssymb}
\usepackage{mathrsfs}
\usepackage{mathtools, cuted}
\usepackage{tabularx, booktabs}
\newcolumntype{C}{>{\centering\arraybackslash}X}
\newcolumntype{R}{>{\raggedleft\arraybackslash}X}
\newcolumntype{S}{>{\raggedleft\arraybackslash\hsize=.5\hsize}X}
\usepackage{latexsym}
\usepackage{url}
\usepackage{xspace}
\usepackage{bm,array}
\usepackage{amsfonts}
\usepackage{enumitem}
\usepackage{cases}
\usepackage{mathtools}
\usepackage{empheq}
\usepackage{bm}
\usepackage{esvect}
\usepackage[noabbrev,capitalize]{cleveref}
\crefname{equation}{equation}{equations}
\crefname{section}{section}{sections}
\crefname{footnote}{footnote}{footnotes}   
\crefname{line}{line}{lines}   
\crefname{assumption}{assumption}{assumptions}

\usepackage{bbm}

\let\frac=\tfrac

\usepackage[compact]{titlesec}

\newcommand{\defn}[1]{\textbf{#1}}
\newcommand{\defeq}{\mathrel{\stackrel{\textnormal{\tiny def}}{=}}}

\newcommand{\Real}{\mathbb{R}}

\newcommand{\Uniform}{\mathrm{Unif}}
\newcommand{\Exp}{\mathrm{Exp}}

\newcommand{\param}{\theta}

\newcommand{\history}{\mathcal{H}}

\newcommand{\es}[2]{#1_{#2}}

\newcommand{\optparens}[1]{\if\relax\detokenize{#1}\relax\else(#1)\fi}

\newcommand{\tree}[2]{%
	\ifthenelse{\equal{#1}{a}}{$\sqrt{#2}$}{%
		\ifthenelse{\equal{#1}{b}}{"Hi."}{}}}

\usepackage{tikz}
\usetikzlibrary{shapes.geometric}

\usetikzlibrary{arrows,decorations.markings}
\usetikzlibrary{arrows}

\usepackage{verbatim}

\title{HYPRO: A Hybridly Normalized Probabilistic Model for Long-Horizon Prediction of Event Sequences}

\author{
  Siqiao Xue, Xiaoming Shi, James Y Zhang\\
  Ant Group\\
  569 Xixi Road\\
  Hangzhou, China\\
  \texttt{siqiao.xsq@alibaba-inc.com}\\
  \texttt{\{peter.sxm,james.z\}@antgroup.com}
  \And
  Hongyuan Mei\\
  Toyota Technological Institute at Chicago\\
  6045 S Kenwood Ave\\
  Chicago, IL 60637\\
  \texttt{hongyuan@ttic.edu}
}

\begin{document}

\maketitle

\begin{abstract}\label{sec:abstract}
    In this paper, we tackle the important yet under-investigated problem of making long-horizon prediction of event sequences. 
    Existing state-of-the-art models do not perform well at this task due to their autoregressive structure. 
    We propose HYPRO, a hybridly normalized probabilistic model that naturally fits this task: its first part is an autoregressive base model that learns to propose predictions; its second part is an energy function that learns to reweight the proposals such that more realistic predictions end up with higher probabilities. 
    We also propose efficient training and inference algorithms for this model. 
    Experiments on multiple real-world datasets demonstrate that our proposed HYPRO model can significantly outperform previous models at making long-horizon predictions of future events. 
    We also conduct a range of ablation studies to investigate the effectiveness of each component of our proposed methods. 
\end{abstract}

\section{Introduction}\label{sec:intro}

Long-horizon prediction of event sequences is essential in various real-world  applied domains: 
\begin{itemize}[leftmargin=*]
\item {\em Healthcare.} Given a patient's symptoms and treatments so far, we would be interested in predicting their future health conditions \emph{over the next several months}, including their prognosis and treatment.

\item {\em Commercial.} Given an online consumer's previous purchases and reviews, we may be interested in predicting what they would buy \emph{over the next several weeks} and plan our advertisement accordingly. 

\item {\em Urban planning.} Having monitored the traffic flow of a town for the past few days, we'd like to predict its future traffic \emph{over the next few hours}, which would be useful for congestion management. 

\item Similar scenarios arise in {\em computer systems}, {\em finance}, {\em dialogue}, {\em music}, etc.
\end{itemize}
Though being important, this task has been under-investigated: the previous work in this research area has been mostly focused on the prediction of the \emph{next single} event (e.g., its time and type). 

In this paper, we show that previous state-of-the-art models suffer at making long-horizon predictions, i.e., predicting \emph{the series of future events over a given time interval}. 
That is because those models are all \emph{autoregressive}: predicting each future event is conditioned on all the previously predicted events; an error can not be corrected after it is made and any error will be cascaded through all the subsequent predictions. 
Problems of the same kind also exist in natural language processing tasks such as generation and machine translation~\citep{ranzato-2016,goyal2021characterizing}.

In this paper, we propose a novel modeling framework that learns to make long-horizon prediction of event sequences. 
Our main technical contributions include: 
\begin{itemize}[leftmargin=*]
    \item {\em A new model.} The key component of our framework is HYPRO, a \underline{hy}bridly normalized neural \underline{pro}babilistic model that combines an autoregressive base model with an energy function: the base model learns to propose plausible predictions; the energy function learns to reweight the proposals.
    Although the proposals are generated autoregressively, the energy function reads each entire completed sequence (i.e., true past events together with predicted future events) and learns to assign higher weights to those which appear more realistic \emph{as a whole}.
    
    Hybrid models have already demonstrated effective at natural language processing such as detecting machine-generated text~\citep{bakhtin2019learning} and improving coherency in text generation~\citep{deng-2018-reb}. 
    We are the first to develop a model of this kind for time-stamped event sequences. Our model can use any autoregressive event model as its base model, and we choose the state-of-the-art continuous-time Transformer architecture~\citep{yang-2022-transformer} as its energy function. 

    \item {\em A family of new training objectives.} Our second contribution is a family of training objectives that can estimate the parameters of our proposed model with low computational cost. Our training methods are based on the principle of noise-contrastive estimation since the log-likelihood of our HYPRO model involves an intractable normalizing constant (due to using energy functions). 
    \item {\em A new efficient inference method.} Another contribution is a normalized importance sampling algorithm, which can efficiently draw the predictions of future events over a given time interval from a trained HYPRO model. 
\end{itemize}

\section{Technical Background}\label{sec:prilim}
\subsection{Formulation: Generative Modeling of Event Sequences}\label{sec:form}
We are given a fixed time interval $[0, T]$ over which an event sequence is observed. Suppose there are $I$ events in the sequence at times $0 < t_1 < \ldots < t_I \leq T$. We denote the sequence as $\es{x}{[0, T]} = (t_1, k_1), \ldots, (t_I, k_I)$ where each $k_i \in \{1, \ldots, K\}$ is a discrete event type. 

Generative models of event sequences are temporal point processes. 
They are \defn{autoregressive}: events are generated from left to right; the probability of $(t_i, k_i)$ depends on the history of events $\es{x}{[0,t_{i})} = (t_1, k_1), \ldots, (t_{i-1}, k_{i-1})$ that were drawn at times $< t_i$.  
They are \defn{locally normalized}: if we use $p_k(t \mid \es{x}{[0,t)})$ to denote the probability that an event of type $k$ occurs over the infinitesimal interval $[t, t + dt)$, then the probability that nothing occurs will be $1 - \sum_{k=1}^{K} p_k(t \mid \es{x}{[0,t)})$. 
Specifically, temporal point processes define functions $\lambda_k$ that determine a finite \defn{intensity} $\lambda_k( t \mid \es{x}{[0,t)}) \geq 0$ for each event type $k$ at each time $t > 0$ such that $p_k(t \mid \es{x}{[0,t)}) = \lambda_k( t \mid \es{x}{[0,t)}) dt$. 
Then the log-likelihood of a temporal point process given the entire event sequence $\es{x}{[0, T]}$ is
\begin{align}
    \sum_{i=1}^{I} \log \lambda_{k_i}( t_i \mid \es{x}{[0,t_i)}) - \int_{t=0}^{T} \sum_{k=1}^{K} \lambda_{k}(t \mid \es{x}{[0,t)}) dt \label{eqn:autologlik}
\end{align}

Popular examples of temporal point processes include Poisson processes~\citep{daley-07-poisson} as well as Hawkes processes~\citep{hawkes-71} and their modern neural versions \citep{du-16-recurrent,mei-17-neuralhawkes,zuo2020transformer,zhang-2020-self,yang-2022-transformer}.

\subsection{Task and Challenge: Long-Horizon Prediction and Cascading Errors}\label{sec:task}
We are interested in predicting the future events over an \emph{extended} time interval $(T, T']$. We call this task \defn{long-horizon prediction} as the boundary $T'$ is so large that (with a high probability) many events will happen over $(T, T']$. 
A principled way to solve this task works as follows: we draw many possible future event sequences over the interval $(T, T']$, and then use this empirical distribution to answer questions such as ``how many events of type $k=3$ will happen over that interval''. 

A serious technical issue arises when we draw each possible future sequence. 
To draw an event sequence from an autoregressive model, we have to repeatedly draw the next event, append it to the history, and then continue to draw the next event conditioned on the new history. 
This process is prone to \defn{cascading errors}: any error in a drawn event is likely to cause all the subsequent draws to differ from what they should be, and such errors will accumulate. 

\subsection{Globally Normalized Models: Hope and Difficulties}\label{sec:global}
An ideal fix of this issue is to develop a \defn{globally normalized model} for event sequences. 
For any time interval $[0, T]$, such a model will give a probability distribution that is normalized over all the possible \emph{full sequences} on $[0, T]$ rather than over all the possible instantaneous subsequences within each $(t, t+dt)$. 
Technically, a globally normalized model assigns to each sequence $\es{x}{[0,T]}$ a score $\exp{\left( -E(\es{x}{[0,T]}) \right)}$ where $E$ is called \defn{energy function}; the normalized probability of $\es{x}{[0,T]}$ is proportional to its score: i.e., $p(\es{x}{[0,T]}) \propto \exp{\left( -E(\es{x}{[0,T]}) \right)}$. 

Had we trained a globally normalized model, we wish to enumerate all the possible $\es{x}{[T,T']}$ for a given $\es{x}{[0,T]}$ and select those which give the highest model probabilities $p(\es{x}{[0,T']})$. Prediction made this way would not suffer cascading errors: the entire $\es{x}{[0,T']}$ was jointly selected and thus the overall compatibility between the events had been considered. 

However, training such a globally normalized probabilistic model involves computing the normalizing constant $\sum \exp{\left( -E(\es{x}{[0,T]}) \right)}$ where the summation $\sum$ is taken over all the possible sequences; it is intractable since there are \emph{infinitely} many sequences. 
What's worse, it is also intractable to exactly sample from such a model; approximate sampling is tractable but expensive. 

\section{HYPRO: A Hybridly Normalized Neural Probabilistic Model}\label{sec:cont_ebm}

We propose {HYPRO}, a hybridly normalized neural probabilistic model that combines a temporal point process and an energy function: it enjoys both the efficiency of autoregressive models and the capacity of globally normalized models. 
Our model normalizes over (sub)sequences: for any given interval $[0, T]$ and its extension $(T, T']$ of interest, the model probability of the sequence $\es{x}{(T,T']}$ is 
\begin{align}
    p_{\text{HYPRO}}\left( \es{x}{(T, T']} \mid \es{x}{[0, T]} \right) = p_{\text{auto}}\left( \es{x}{(T, T']} \mid \es{x}{[0, T]} \right) \frac{\exp\left(- E_{\param}(\es{x}{[0, T']}) \right)}{Z_{\param} \left( \es{x}{[0, T]} \right)}
    \label{eqn:reb_tpp}
\end{align}
where $p_{\text{auto}}$ is the probability under the chosen temporal point process and $E_{\param}$ is an energy function with parameters $\param$. 
The normalizing constant sums over all the possible continuations $\es{x}{(T, T']}$ for a given prefix $\es{x}{[0, T]}$: $Z_{\param} \left( \es{x}{[0, T]} \right) \defeq \sum_{\es{x}{(T, T']}} p_{\text{auto}}\left( \es{x}{(T, T']} \mid \es{x}{[0, T]} \right) \exp\left(- E_{\param}(\es{x}{[0, T']}) \right)$. 

The key advantage of our model over autoregressive models is that: the energy function $E_{\param}$ is able to pick up the global features that may have been missed by the autoregressive base model $p_{\text{auto}}$; intuitively, the energy function fits the residuals that are not captured by the autoregressive model. 

Our model is general: in principle, $p_{\text{auto}}$ can be any autoregressive model including those mentioned in \cref{sec:form} and $E_{\param}$ can be any function that is able to encode an event sequence to a real number. 
In \cref{sec:exp}, we will introduce a couple of specific $p_{\text{auto}}$ and $E_{\param}$ and experiment with them. 

In this section, we focus on the training method and inference algorithm. 

\subsection{Training Objectives} \label{sec:training}
Training our full model $p_{\text{HYPRO}}$ is to learn the parameters of the autoregressive model $p_{\text{auto}}$ as well as those of the energy function $E_{\param}$. 
Maximum likelihood estimation (MLE) is undesirable: the objective would be $\log p_{\text{HYPRO}}\left( \es{x}{(T, T']} \mid \es{x}{[0, T]} \right) = \log p_{\text{auto}}\left( \es{x}{(T, T']} \mid \es{x}{[0, T]} \right) - E_{\param}(\es{x}{[0, T']}) - \log Z_{\param} \left( \es{x}{[0, T]} \right)$ where the normalizing constant $Z_{\param} \left( \es{x}{[0, T]} \right)$ is known to be \emph{uncomputable} and \emph{inapproximable} for a large variety of reasonably expressive functions $E_{\param}$~\citep{lin2022on}. 

We propose a training method that works around this normalizing constant. 
We first train $p_{\text{auto}}$ just like how previous work trained temporal point processes.\footnote{It can be done by either maximum likelihood estimation or noise-contrastive estimation: for the former, read \citet{daley-07-poisson}; for the latter, read \citet{mei-2020-nce} which also has an in-depth discussion about the theoretical connections between these two parameter estimation principles.} 
Then we use the trained $p_{\text{auto}}$ as a noise distribution and learn the parameters $\param$ of $E_{\param}$ by noise-contrastive estimation (NCE). 
Precisely, we sample $N$ noise sequences $\es{x}{[T, T']}^{(1)}, \ldots, \es{x}{[T, T']}^{(N)}$, compute the ``energy'' $E_{\param}(\es{x}{[0, T']}^{(n)})$ for each \emph{completed} sequence $\es{x}{[0, T']}^{(n)}$, and then plug those energies into one of the following training objectives.

Note that all the completed sequences $\es{x}{[0, T']}^{(n)}$ share the same observed prefix $\es{x}{[0, T]}$. 

\paragraph{Binary-NCE Objective.}
We train a binary classifier based on the energy function $E_{\param}$ to discriminate the true event sequence---denoted as $\es{x}{[0, T']}^{(0)}$---against the noise sequences by maximizing  
\begin{align}
      J_{\text{binary}} = \log \sigma \left( -E_{\param}(\es{x}{[0, T']}^{(0)}) \right) + \sum_{n=1}^{N} \log \sigma \left( E_{\param}(\es{x}{[0, T']}^{(n)}))\right)
    \label{eqn:bin_entropy_loss}
\end{align}
where $\sigma(u) = \frac{1}{1+\exp(-u)}$ is the sigmoid function. 
By maximizing this objective, we are essentially pushing our energy function $E_{\param}$ such that the observed sequences have \emph{low} energy but the noise sequences have \emph{high} energy. 
As a result, the observed sequences will be \emph{more probable} under our full model $p_{\text{HYPRO}}$ while the noise sequences will be \emph{less probable}: see \cref{eqn:reb_tpp}. 

Theoretical guarantees of general Binary-NCE can be found in \citet{gutmann-10-nce}. 
For general conditional probability models like ours, Binary-NCE implicitly assumes self-normalization~\citep{mnih-12-nce,ma-18-nce}: i.e., $Z_{\param}\left( \es{x}{[0, T]} \right) = 1$ is satisfied.

This type of training objective has been used to train a hybridly normalized text generation model by \citet{deng-2018-reb}; see \cref{sec:related} for more discussion about its relations with our work. 

\paragraph{Multi-NCE Objective.}
Another option is to use Multi-NCE objective\footnote{It was named as Ranking-NCE by \citet{ma-18-nce}, but we think Multi-NCE is a more appropriate name since it constructs a multi-class classifier over one correct answer and multiple incorrect answers.}, which means we maximize
\begin{align}
      J_{\text{multi}} =  -E_{\param}(\es{x}{[0, T']}^{(0)}) - \log \sum_{n=0}^{N} \exp\left( - E_{\param}(\es{x}{[0, T']}^{(n)}))\right)
    \label{eqn:multi_entropy_loss}
\end{align}
By maximizing this objective, we are pushing our energy function $E_{\param}$ such that each observed sequence has \emph{relatively lower} energy than the noise sequences sharing the same observed prefix. 
In contrast, $J_{\text{binary}}$ attempts to make energies \emph{absolutely} low (for observed data) or high (for noise data) without considering whether they share prefixes. 
This effect is analyzed in Analysis-III of \cref{sec:analysis}. 

This $J_{\text{multi}}$ objective also enjoys better statistical properties than $J_{\text{binary}}$ since it doesn't assume self-normalization: the normalizing constant $Z_{\param}\left( \es{x}{[0, T]} \right)$ is neatly cancelled out in its derivation; see \cref{app:nce} for a full derivation of both Binary-NCE and Multi-NCE. 

Theoretical guarantees of Multi-NCE for discrete-time models were established by \citet{ma-18-nce}; \citet{mei-2020-nce} generalized them to temporal point processes. 

\paragraph{Considering Distances Between Sequences.}
Previous work~\citep{lecun-2006,bakhtin2019learning} reported that energy functions may be better learned if the distances between samples are considered. 
This has inspired us to design a regularization term that enforces such consideration. 

Suppose that we can measure a well-defined ``distance'' between the true sequence $\es{x}{[0, T']}^{(0)}$ and any noise sequence $\es{x}{[0, T']}^{(n)}$; we denote it as $d(n)$. 
We encourage the energy of each noise sequence to be higher than that of the observed sequence by a margin; that is, we propose the following regularization:\looseness=-1
% term to our objectives: 
\begin{align}
    %J_{\text{binary}} + \Omega
    %{ \text{ or } }
    %J_{\text{multi}} + \Omega
    %{ \text{ where } }
    \Omega = \sum_{n=1}^{N} \max \left( 0, \beta d(n) + E_{\param}(\es{x}{[0, T']}^{(0)}) - E_{\param}(\es{x}{[0, T']}^{(n)}) \right)\label{eqn:reg}
\end{align}
where $\beta > 0$ is a hyperparameter that we tune on the held-out development data. 
With this regularization, the energies of the sequences with larger distances will be pulled farther apart: this will help discriminate not only between the observed sequence and the noise sequences, but also between the noise sequences themselves, thus making the energy function $E_{\param}$ more informed. 

This method is general so the distance $d$ can be any appropriately defined metric. In \cref{sec:exp}, we will experiment with an {optimal transport distance} specifically designed for event sequences.

Note that the distance $d$ in the regularization may be the final test metric. 
In that case, our method is directly optimizing for the final evaluation score.

\paragraph{Generating Noise Sequences.} 
Generating event sequences from an autoregressive temporal point process has been well-studied in previous literature.
The standard way is to call the \defn{thinning algorithm}~\citep{lewis-79-sim,liniger-09-hawkes}. 
The full recipe for our setting is in \cref{alg:noise_sampling}.
\begin{algorithm}[tb]
    \caption{Generating Noise Sequences.}\label{alg:noise_sampling}
    \begin{algorithmic}[1]
		\INPUT an event sequence $\es{x}{[0,T]}$ over the given interval $[0,T]$ and an interval $(T, T']$ of interest; \newline
		trained autoregressive model $p_{\text{auto}}$ and number of noise samples $N$ 
		\OUTPUT a collection of noise sequences
		\Procedure{DrawNoise}{$\es{x}{[0,T]}, T', p_{\text{auto}}, N$}
		\For{$n=1$ {\bfseries to} $N$}
		    \LineComment{use the thinning algorithm to draw each noise sequences from the autoregressive model $p_{\text{auto}}$}
		    \LineComment{in particular, call the method in \cref{alg:thinning} that is described in \cref{app:thinning}}
		    \State $\es{x}{(T,T']}^{(n)} \gets$ \textsc{Thinning}($\es{x}{[0,T]}, T', p_{\text{auto}}$)
		\EndFor
		\State \textbf{return} $\es{x}{(T,T']}^{(1)}, \ldots, \es{x}{(T,T']}^{(N)}$
		\EndProcedure
	\end{algorithmic}
\end{algorithm}
\subsection{Inference Algorithm}\label{sec:infer}
Inference involves drawing future sequences $\es{x}{(T,T']}$ from the trained full model $p_{\text{HYPRO}}$; due to the uncomputability of the normalizing constant $Z(\es{x}{[0,T]})$, exact sampling is intractable.

We propose a \defn{normalized importance sampling} method to approximately draw $\es{x}{(T,T']}$ from $p_{\text{HYPRO}}$; it is shown in \cref{alg:joint_sampling}. 
We first use the trained $p_{\text{auto}}$ to be our proposal distribution and call the thinning algorithm (\cref{alg:noise_sampling}) to draw proposals $\es{x}{[T, T']}^{\langle1\rangle}, \ldots, \es{x}{[T, T']}^{\langle M\rangle}$.
Then we reweight those proposals with the \emph{normalized} weights $w^{\langle m \rangle}$ that are defined as
\begin{align}
    w^{\langle m \rangle} 
    \defeq \frac{p_{\text{HYPRO}}(\es{x}{[T, T']}^{\langle m\rangle}) / p_{\text{auto}}(\es{x}{[T, T']}^{\langle m\rangle})}{\sum_{m'=1}^{M} p_{\text{HYPRO}}(\es{x}{[T, T']}^{\langle m'\rangle}) / p_{\text{auto}}(\es{x}{[T, T']}^{\langle m'\rangle})}
    = \frac{\exp\left(-E_{\param}(\es{x}{[0,T']}^{\langle m \rangle})\right)}{\sum_{m'=1}^{M} \exp\left(-E_{\param}(\es{x}{[0,T']}^{\langle m' \rangle})\right) }
\end{align}
This collection of weighted proposals is used for the long-horizon prediction over the interval $(T, T']$: 
if we want the most probable sequence, we return the $\es{x}{[T, T']}^{\langle m\rangle}$ with the largest weight $w^{\langle m \rangle}$; 
if we want a minimum Bayes risk prediction (for a specific risk metric), we can use existing methods (e.g., the consensus decoding method in \citet{mei-19-smoothing}) to compose those weighted samples into a single sequence that minimizes the risk. In our experiments (\cref{sec:exp}), we used the most probable sequence.\looseness=-1

Note that our sampling method is \emph{biased} since the weights $w^{\langle m \rangle}$ are \emph{normalized}. 
Unbiased sampling in our setting is intractable since that will need our weights to be unnormalized: i.e., \mbox{$w \defeq p_{\text{HYPRO}} / p_{\text{auto}} = \exp\left(-E_{\param}\right) / Z$} which circles back to the problem of $Z$'s uncomputability. 
Experimental results in \cref{sec:exp} show that our method indeed works well in practice despite that it is biased.\looseness=-1
\begin{algorithm}
    \caption{Normalized Importance Sampling for Long-Horizon Prediction.}\label{alg:joint_sampling}
    \begin{algorithmic}[1]
		\INPUT an event sequence $\es{x}{[0,T]}$ over the given interval $[0,T]$ and an interval $(T, T']$ of interest; \newline
		trained autoregressive model $p_{\text{auto}}$ and engergy function $E_{\param}$, number of proposals $M$ 
		\OUTPUT a collection of weighted proposals
		\Procedure{NIS}{$\es{x}{[0,T]}, T', p_{\text{auto}}, E_{\param}, M$}
		\LineComment{use normalized importance sampling to approximately draw $M$ proposals from $p_{\text{HYPRO}}$}
		\State $\es{x}{[T, T']}^{\langle1\rangle}, \ldots, \es{x}{[T, T']}^{\langle M\rangle} \gets$ \textsc{DrawNoise}($\es{x}{[0,T]}, T', p_{\text{auto}}, M$) \Comment{see \cref{alg:noise_sampling}}
		\State construct completed sequences $\es{x}{[0, T']}^{\langle1\rangle}, \ldots, \es{x}{[0, T']}^{\langle M\rangle}$ by appending each $\es{x}{[T, T']}^{\langle m\rangle}$ to $\es{x}{[0,T]}$
		\State compute the exponential of minus energy $e^{\langle m\rangle} = \exp\left(-E_{\param}(\es{x}{[0, T']}^{\langle m\rangle})\right)$ for each proposal
		\State compute the normalized weights $w^{\langle m\rangle} = e^{\langle m\rangle} / \sum_{m'=1}^{M} e^{\langle m'\rangle}$
		\State {\bfseries return} $( w^{\langle 1\rangle}, \es{x}{[T, T']}^{\langle1\rangle} ), \ldots, ( w^{\langle M\rangle}, \es{x}{[T, T']}^{\langle M\rangle} )$ \Comment{return the collection of weighted proposals}
		\EndProcedure
	\end{algorithmic}
\end{algorithm}

\section{Related work}\label{sec:related}

Over the recent years, various neural temporal point processes have been proposed. 
Many of them are built on recurrent neural networks, or LSTMs~\citep{hochreiter-97-lstm}; they include \citet{du-16-recurrent,mei-17-neuralhawkes,xiao-17-joint,xiao-17-modeling,omi-19-fully,shchur-20-intensity,mei-2020-datalog,boyd-20-vae}. 
Some others use Transformer architectures~\citep{vaswani-2017-transformer,radford-2019-gpt}: in particular, \citet{zuo2020transformer,zhang-2020-self,enguehard-2020-neural,sharma-2021-identifying,zhu-2020-deep,yang-2022-transformer}. 
All these models all autoregressive: they define the probability distribution over event sequences in terms of a sequence of locally-normalized conditional distributions over events given their histories. 

Energy-based models, which have a long history in machine learning~\citep{hop-1982,hinton-2002,lecun-2006,ranzato-2007-eb,ng-learn-dem-2011,xie-2019}, define the distribution over sequences in a different way: they use energy functions to summarize each possible sequence into a scalar (called energy) and define the unnormalized probability of each sequence in terms of its energy, then the probability distribution is normalized across all sequences; thus, they are also called globally normalized models.
Globally normalized models are a strict generalization of locally normalized models~\citep{lin-et-al-2021-naacl}: all the locally normalized models are globally normalized; but the converse is not true. 
Moreover, energy functions are good at capturing global features and structures~\citep{pang-2021,du-ebm-2019,brakel-imputation-2013}. 
However, the normalizing constants of globally normalized models are often uncomputable~\citep{lin2022on}. 
Existing work that is most similar to ours is the energy-based text generation models of \citet{bakhtin2019learning} and \citet{deng-2018-reb} that train energy functions to reweight the outputs generated by pretrained autoregressive models. 
The differences are: we work on different kinds of sequential data (continuous-time event sequences vs.\@ discrete-time natural language sentences), and thus the architectures of our autoregressive model and energy function are different from theirs; additionally, we explored a wider range of training objectives (e.g., Multi-NCE) than they did.

The task of long-horizon prediction has drawn much attention in several machine learning areas such as regular time series analysis~\citep{yu2019long,le2019shape}, natural language processing~\citep{guo2018long,guan2021long}, and speech modeling~\citep{oord2016wavenet}.
\citet{Deshpande_2021} is the best-performing to-date in long-horizon prediction of event sequences: they adopt a hierarchical architecture similar to ours and use a ranking objective based on the counts of the events. 
Their method can be regarded as a special case of our framework (if we let our energy function read the counts of the events), and our method works better in practice (see \cref{sec:exp}). 

\section{Experiments}\label{sec:exp}

We implemented our methods with PyTorch~\citep{paszke-17-pytorch}. 
Our code can be found at {\small \url{https://github.com/alipay/hypro_tpp}} and {\small \url{https://github.com/iLampard/hypro_tpp}}.
Implementation details can be found in \cref{app:imp}.

\subsection{Experimental Setup}
Given a train set of sequences, we use the full sequences to train the autoregressive model $p_{\text{auto}}$ by maximizing \cref{eqn:autologlik}. To train the energy function $E_{\param}$, we need to split each sequence into a prefix $\es{x}{[0,T]}$ and a continuation $\es{x}{(T,T']}$: we choose $T$ and $T'$ such that there are 20 event tokens within $(T, T']$ on average. 
During testing, for each prefix $\es{x}{[0,T]}$, we draw 20 weighted samples (\cref{alg:joint_sampling}) and choose the highest-weighted one as our prediction $\es{\hat{x}}{(T,T']}$. 
We evaluate our predictions by: 
\begin{itemize}[leftmargin=*]
    \item The root of mean square error (RMSE) of the number of the tokens of each event type: for each type $k$, we count the number of type-$k$ tokens in the true continuation---denoted as $C_{k}$---as well as that in the prediction---denoted as $\hat{C}_{k}$; then the mean square error is $\sqrt{\frac{1}{K}\sum_{k=1}^{K} \left( C_k - \hat{C}_k \right)^2}$.
    \item The optimal transport distance (OTD) between event sequences defined by \citet{mei-19-smoothing}: for any given prefix $\es{x}{[0,T]}$, the distance is defined as the minimal cost of editing the prediction $\es{\hat{x}}{(T,T']}$ (by inserting or deleting events, changing their occurrence times, and changing their types) such that it becomes exactly the same as the true continuation $\es{x}{(T,T']}$.
\end{itemize}

We did experiments on two real-world datasets (see \cref{app:data_stat} for dataset details):
\begin{itemize}[leftmargin=*]
\item {\bfseries Taobao \textnormal{\citep{tianchi-taobao-2018}}.} This public dataset was created and released for the 2018 Tianchi Big Data Competition. It contains time-stamped behavior records (e.g., browsing, purchasing) of anonymized users on the online shopping platform Taobao from November 25 through December 03, 2017. 
Each category group (e.g., men's clothing) is an event type, and we have $K=17$ event types. 
We use the browsing sequences of the most active 2000 users; each user has a sequence. Then we randomly sampled disjoint train, dev and test sets with $1300$, $200$ and $500$ sequences. 
The time unit is $3$ hours; the average inter-arrival time is $0.06$ (i.e., $0.18$ hour), and we choose the prediction horizon $T'-T$ to be $1.5$ that approximately covers $20$ event tokens.

\item {\bfseries Taxi \textnormal{\citep{whong-14-taxi}}.} This dataset tracks the time-stamped taxi pick-up and drop-off events across the five boroughs of the New York city; each (borough, pick-up or drop-off) combination defines an event type, so there are $K=10$ event types in total. We work on a randomly sampled subset of $2000$ drivers and each driver has a sequence. We randomly sampled disjoint train, dev and test sets with $1400$, $200$ and $400$ sequences.
The time unit is $1$ hour; the average inter-arrival time is $0.22$, and we set the prediction horizon to be $4.5$ that approximately covers $20$ event tokens.

\item {\bfseries StackOverflow \textnormal{\citep{snapnets}}.} This dataset has two years of user awards on a question-answering website: each user received a sequence of badges and there are $K=22$ different kinds of badges in total. 
We randomly sampled disjoint train, dev and test sets with $1400,400$ and $400$ sequences from the dataset. 
The time unit is $11$ days; the average inter-arrival time is $0.95$ and we set the prediction horizon to be $20$ that approximately covers $20$ event tokens.

\end{itemize}

We choose two strong autoregressive models as our base model $p_{\text{auto}}$:
\begin{itemize}[leftmargin=*]
    \item {\bfseries Neural Hawkes process (NHP) \textnormal{\citep{mei-17-neuralhawkes}}.} It is an LSTM-based autoregressive model that has demonstrated effective at modeling event sequences in various domains. 
    \item {\bfseries Attentative neural Hawkes process (AttNHP) \textnormal{\citep{yang-2022-transformer}}.} It is an attention-based autoregressive model---like Transformer language model~\citep{vaswani-2017-transformer,radford-2019-gpt}---whose performance is comparable to or better than that of the NHP as well as other attention-based models~\citep{zuo2020transformer,zhang-2020-self}.
\end{itemize}

For the energy function $E_{\param}$, we adapt the \defn{continuous-time Transformer module} of the AttNHP model: 
the Transformer module embeds the given sequence of events $x$ into a fixed-dimensional vector (see section-2 of \citet{yang-2022-transformer} for details), which is then mapped to a scalar $\in \Real$ via a multi-layer perceptron (MLP); that scalar is the energy value $E_{\param}(x)$. 

We first train the two base models NHP and AttNHP; they are also used as the baseline methods that we will compare to. 
To speed up energy function training, we use the pretrained weights of the AttNHP to initialize the Transformer part of the energy function; this trick was also used in \citet{deng-2018-reb} to bootstrap the energy functions for text generation models. 
As the full model $p_{\text{HYPRO}}$ has significantly more parameters than the base model $p_{\text{auto}}$, we also trained larger NHP and AttNHP with comparable amounts of parameters as extra baselines. 
Additionally, we also compare to the \defn{DualTPP} model of \citet{Deshpande_2021}. 
Details about model parameters are in \cref{tab:model_params} of \cref{app:training_details}.

\subsection{Results and Analysis}\label{sec:results}\label{sec:analysis}

\begin{figure*}[t]
	\begin{center}
	    \begin{subfigure}[t]{0.3\linewidth}
                \includegraphics[width=0.99\linewidth]{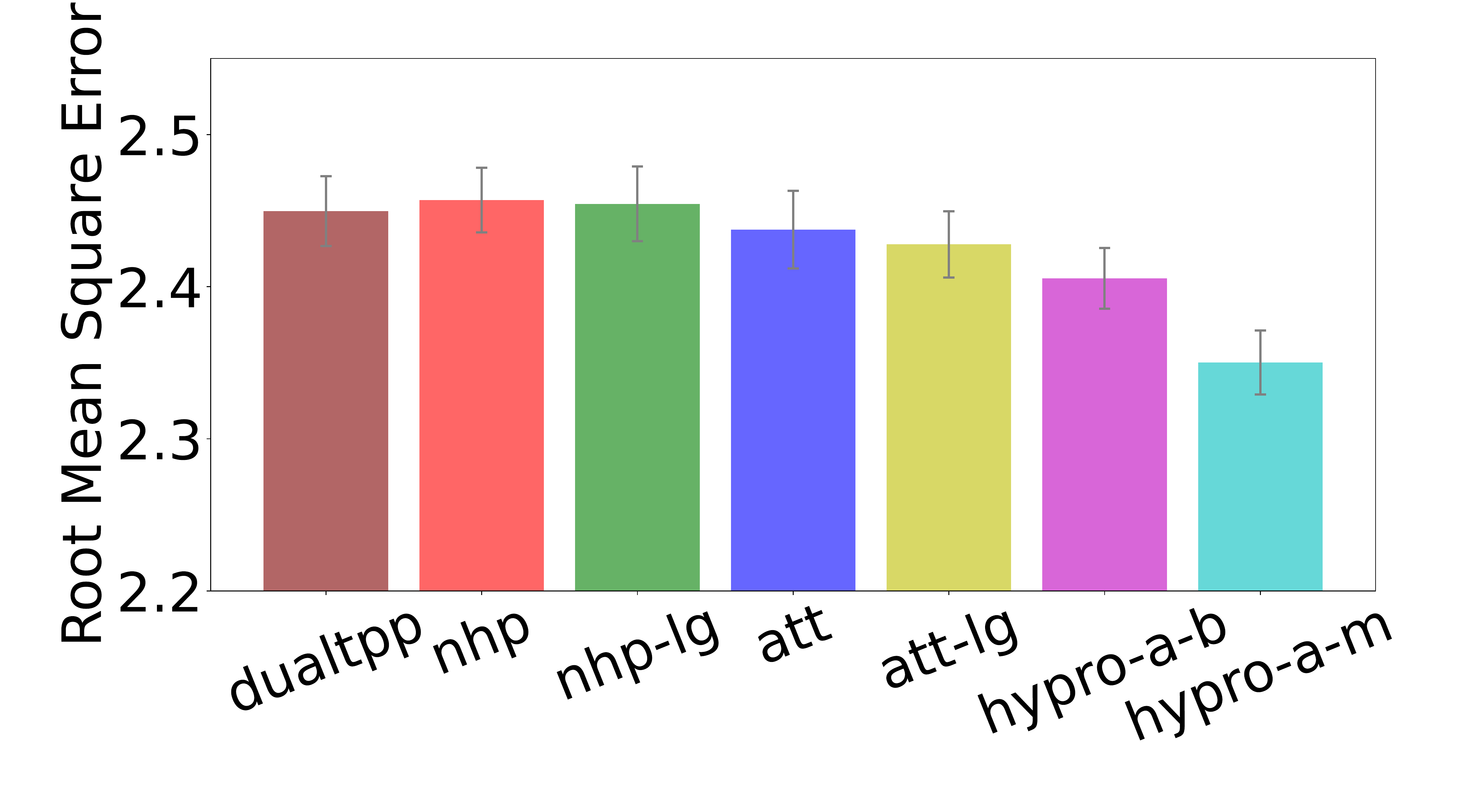}
    		\vfill
                \includegraphics[width=0.99\linewidth]{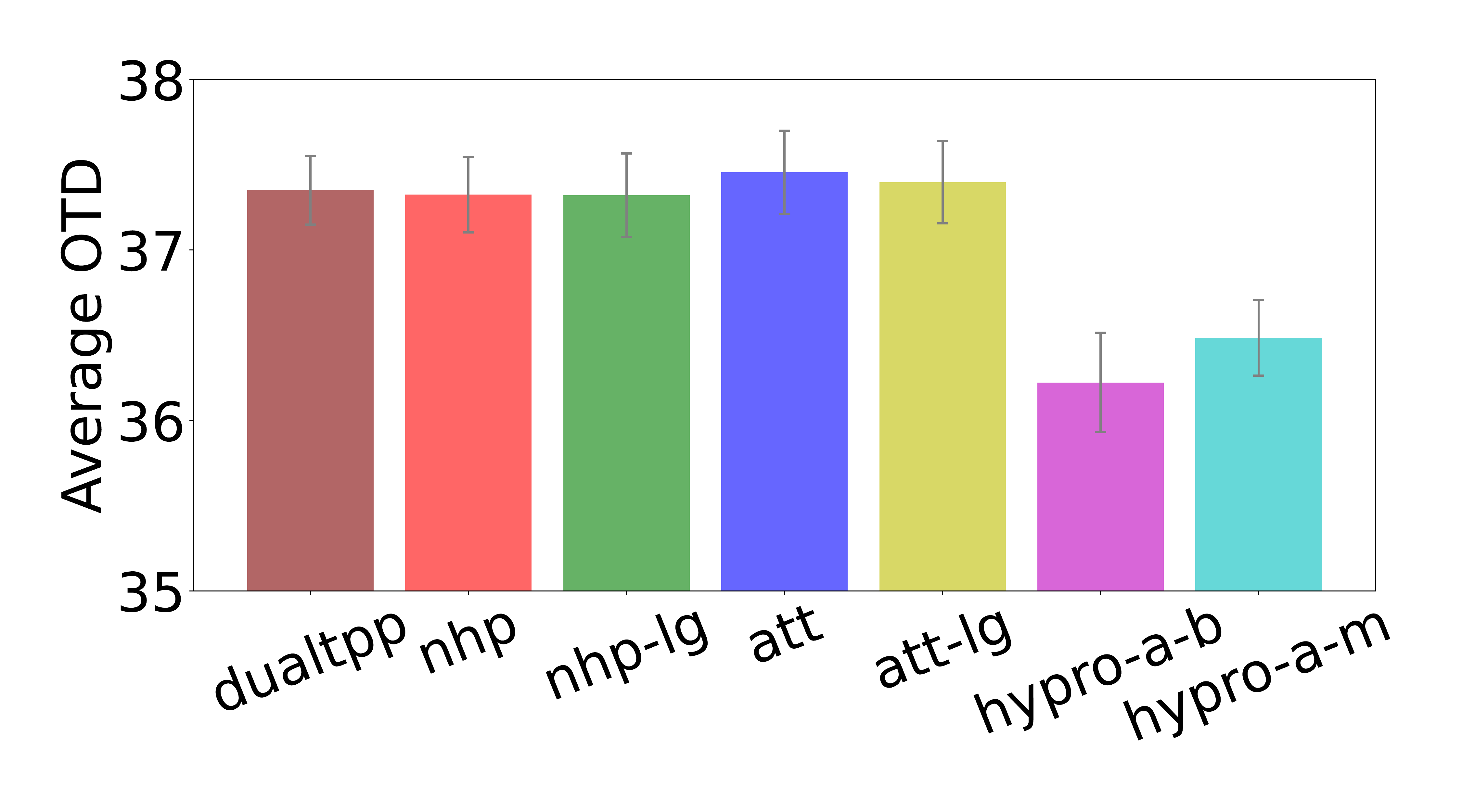}
		    \vspace{-16pt}
			\caption{Taobao Data}\label{fig:main_taobao}
		\end{subfigure}
		\hfill
            \begin{subfigure}[t]{0.3\linewidth}
			\includegraphics[width=0.99\linewidth]{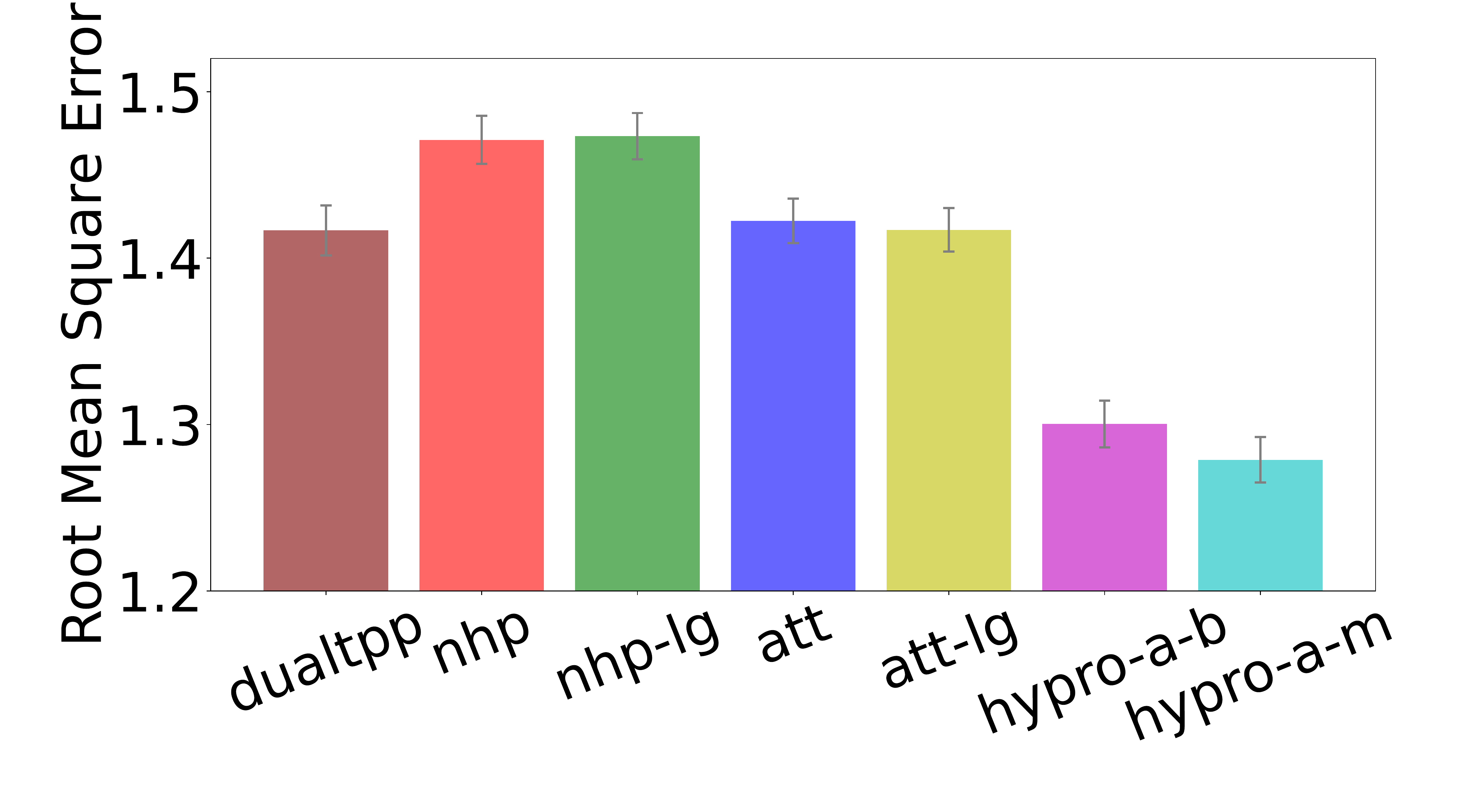}
			\vfill
			\includegraphics[width=0.99\linewidth]{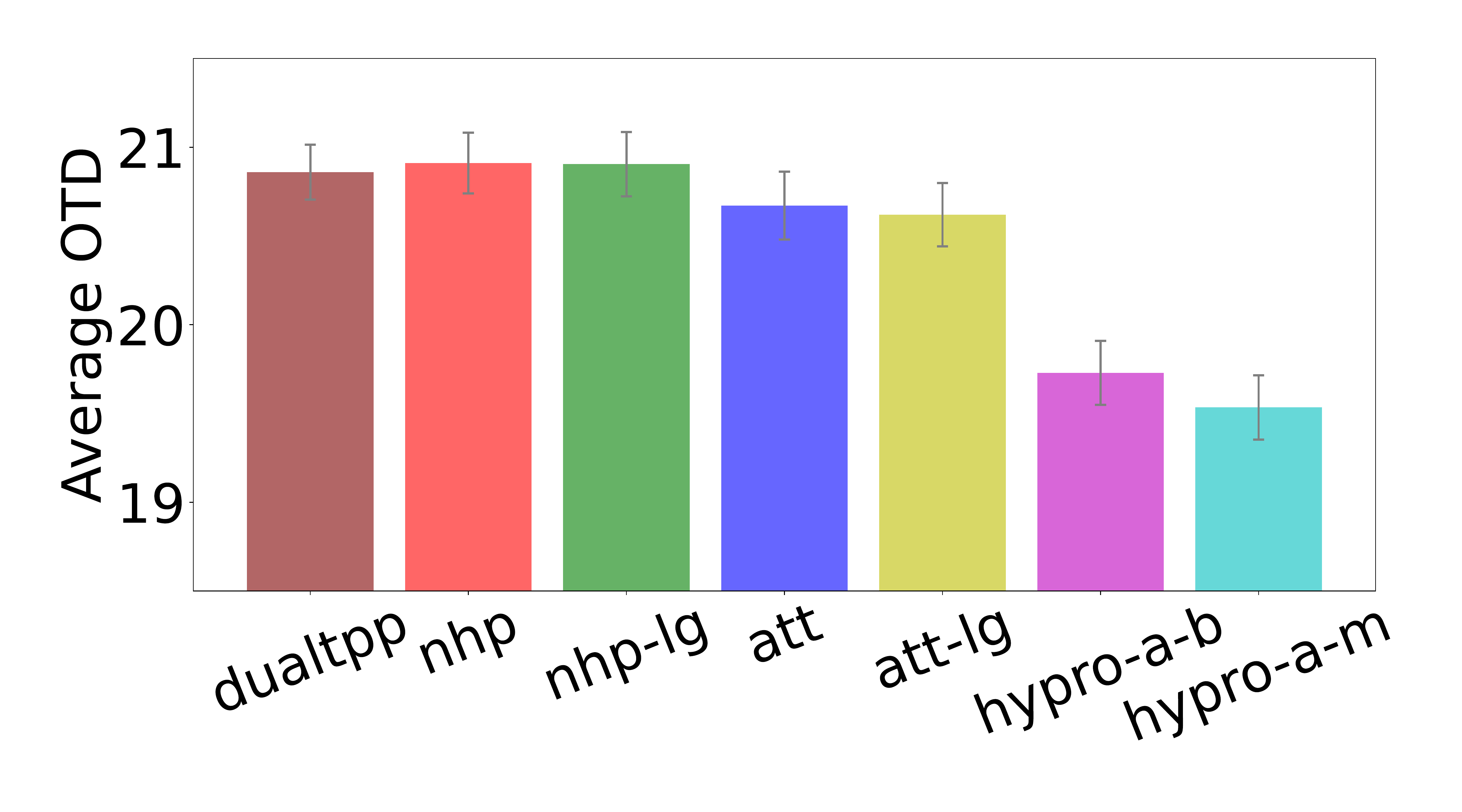}
			\vspace{-16pt}
			\caption{Taxi Data}\label{fig:main_taxi}
		\end{subfigure}
            \hfill
            \begin{subfigure}[t]{0.3\linewidth}
			\includegraphics[width=0.99\linewidth]{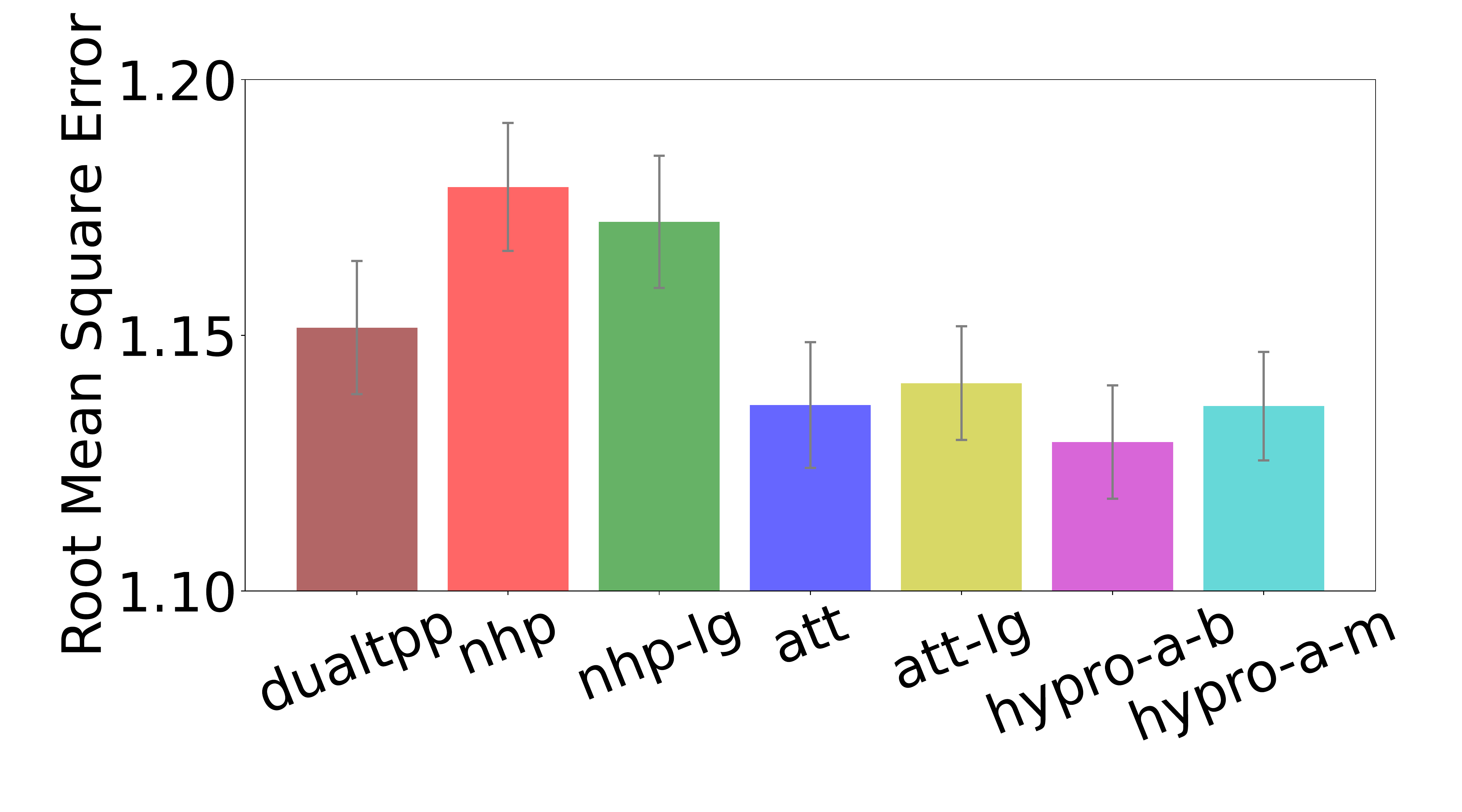}
			\vfill
			\includegraphics[width=0.99\linewidth]{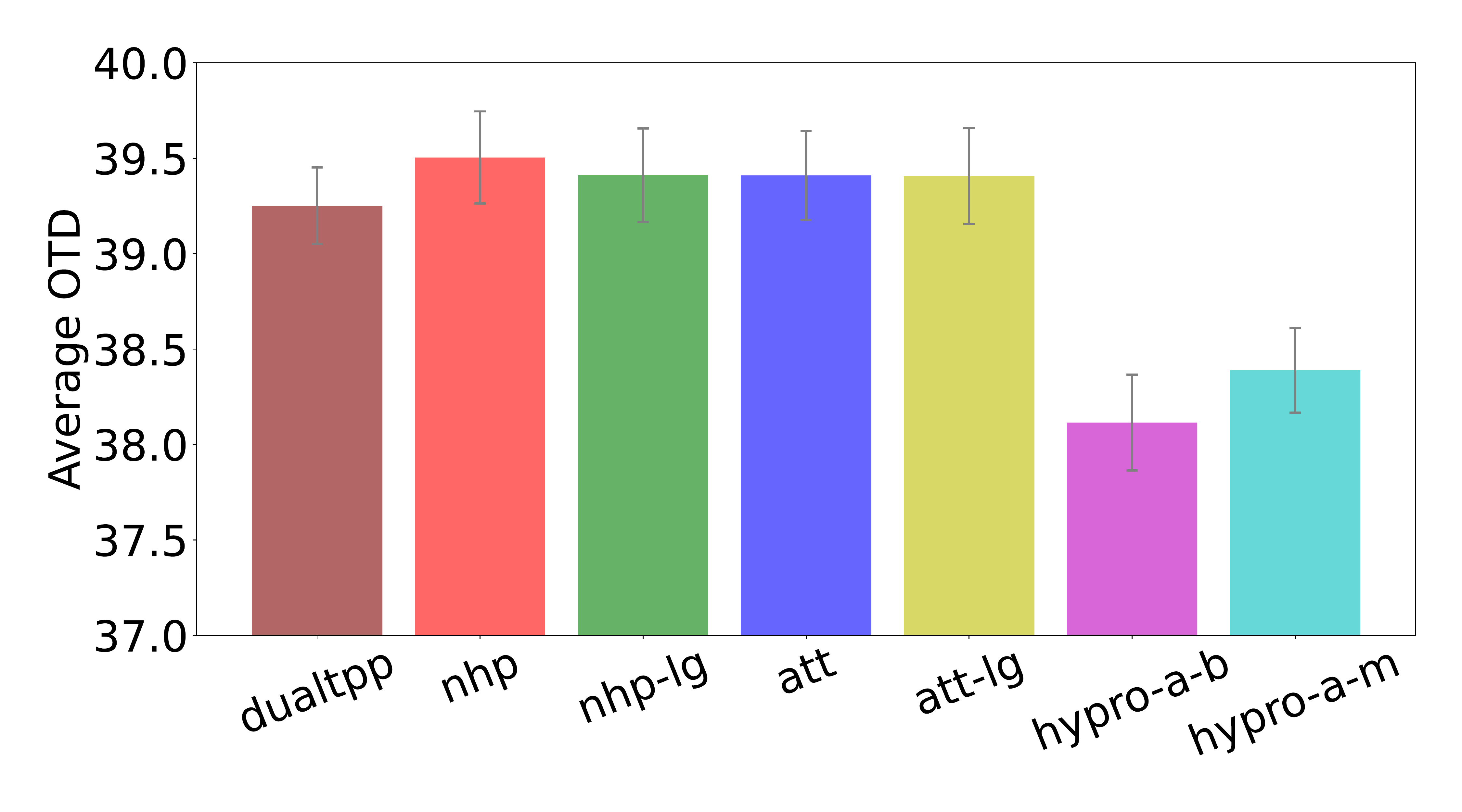}
			\vspace{-16pt}
			\caption{StackOverflow Data}\label{fig:main_so}
		\end{subfigure}
		\vspace{-4pt}
		\caption{Performance of all the methods on Taobao (\ref{fig:main_taobao}), Taxi (\ref{fig:main_taxi}) and StackOverflow (\ref{fig:main_so}) datasets, measured by RMSE (up) and OTD (down). In each figure, the models from left to right are: DualTPP (dualtpp); NHP (nhp); NHP with more parameters (nhp-lg); AttNHP (att); AttNHP with more parameters (att-lg); our HYPRO with Transformer energy function trained via Binary-NCE (hypro-a-b) and Multi-NCE (hypro-a-m).}\label{fig:main_results}
	\end{center}
\end{figure*}

The main results are shown in \cref{fig:main_results}.
The OTD depends on the hyperparameter $C_{\text{del}}$, which is the cost of deleting or adding an event token of any type, so we used a range of values of $C_{\text{del}}$ and report the averaged OTD in \cref{fig:main_results}; OTD for each specific $C_{\text{del}}$ can be found in \cref{app:exp_results}.
As we can see, NHP and AttNHP work the worst in most cases.
DualTPP doesn't seem to outperform these autoregressive baselines even though it learns to rerank sequences based on their macro statistics; we believe that it is because DualTPP's base autoregressive model is not as powerful as the state-of-the-art NHP and AttNHP and using macro statistics doesn't help enough. 
Our HYPRO method works significantly better than these baselines. 

\paragraph{Analysis-I: Does the Cascading Error Exist?} 
Handling cascading errors is a key motivation for our framework. On the Taobao dataset, we empirically confirmed that this issue indeed exists. 

We first investigate whether the event type prediction errors are cascaded through the subsequent events. 
We grouped the sequences based on how early the first event type prediction error was made and then compared the event type prediction error rate on the subsequent events: 
\begin{itemize}[leftmargin=*]
    \item when the first error is made on the first event token, the base AttNHP model has a 71.28\% error rate on the subsequent events and our hybrid model has a much lower 66.99\% error rate. 
    \item when the first error is made on the fifth event token, the base AttNHP model has a 58.88\% error rate on the subsequent events and our hybrid model has a much lower 51.89\% error rate. 
    \item when the first error is made on the tenth event token, the base AttNHP model has a 44.48\% error rate on the subsequent events and our hybrid model has a much lower 35.58\% error rate. 
\end{itemize}
Obviously, when mistakes are made earlier in the sequences, we tend to end up with a higher error rate on the subsequent predictions; that means event type prediction errors are indeed cascaded through the subsequent predictions. Moreover, in each group, our hybrid model enjoys a lower prediction error; that means it indeed helps mitigate this issue.

We then investigate whether the event time prediction errors are cascaded. For this, we performed a linear regression: the independent variable $x$ is the absolute error of the prediction on the time of the first event token; the dependent variable $y$ is the averaged absolute error of the prediction on the time of the subsequent event tokens. Our fitted linear model is $y=0.7965x+0.3219$ where the p-value of the coefficient of $x$ is $\approx 0.0001 < 0.01$. It means that the time prediction errors are also cascaded through the subsequent predictions.

\paragraph{Analysis-II: Energy Function or Just More Parameters?}
The larger NHP and AttNHP have almost the same numbers of parameters with HYPRO, but their performance is only comparable to the smaller NHP and AttNHP. That is to say, simply increasing the number of parameters in an autoregressive model will not achieve the performance of using an energy function. 

To further verify the usefulness of the energy function, we also compared our method with another baseline method that ranks the completed sequences based on their probabilities under the base model, from which the continuations were drawn. This baseline is similar to our proposed HYPRO framework but its scorer is the base model itself.
In our experiments, this baseline method is not better than our method; details can be found in \cref{app:newbaseline}. 

Overall, we can conclude that the energy function $E_{\param}$ is essential to the success of HYPRO.

\begin{figure*}[t]
	\begin{center}
	    \begin{subfigure}[t]{0.99\linewidth}
            \includegraphics[width=0.5\linewidth]{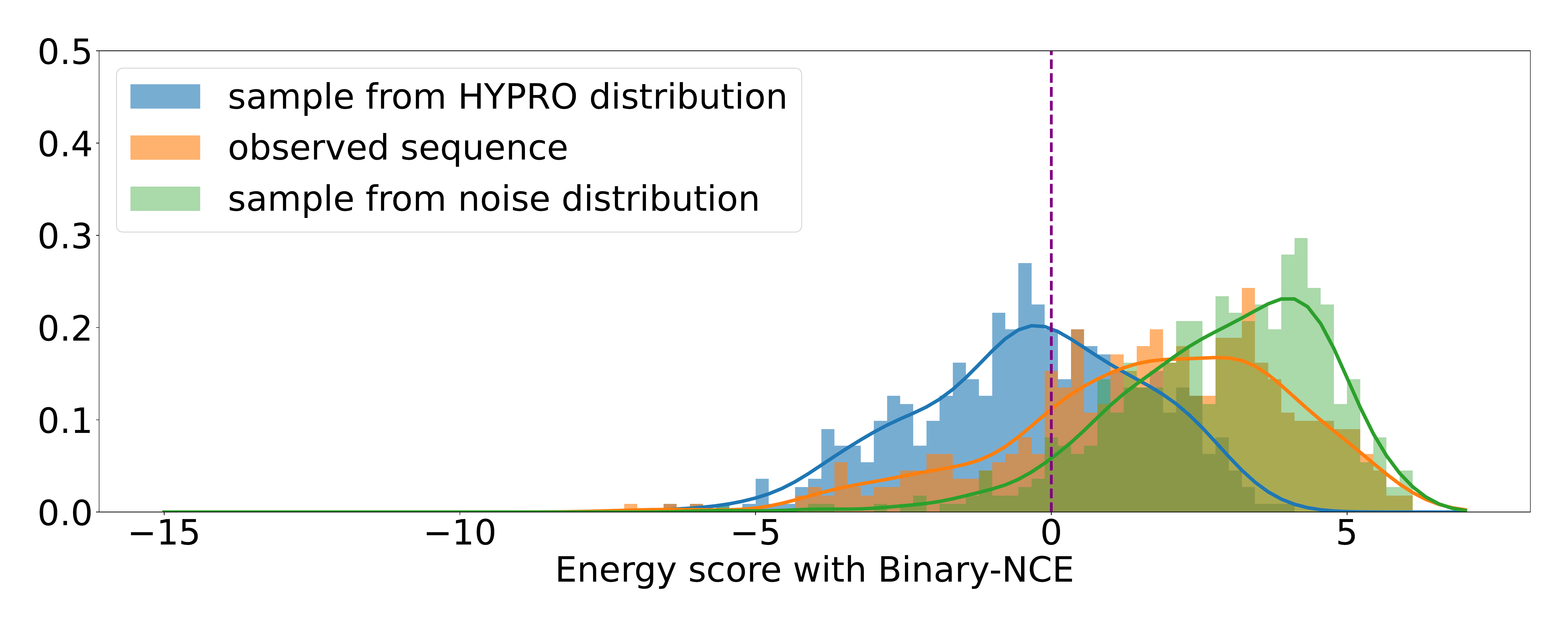}
			~
			\includegraphics[width=0.5\linewidth]{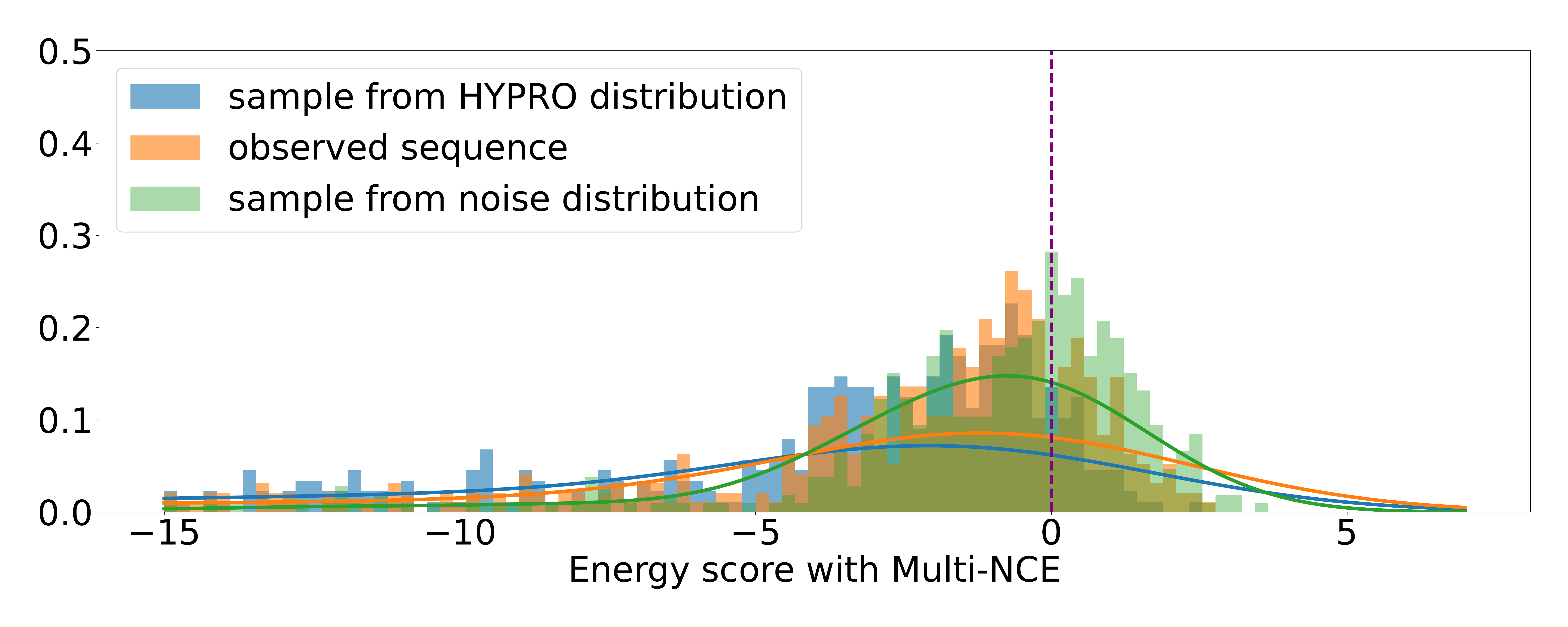}
			\vspace{-20pt}
			\caption{Without the using distance-based regularization of \cref{eqn:reg}.}\label{fig:unreg_energy_distribution}
		\end{subfigure}
		
		\begin{subfigure}[t]{0.99\linewidth}
			\includegraphics[width=0.5\linewidth]{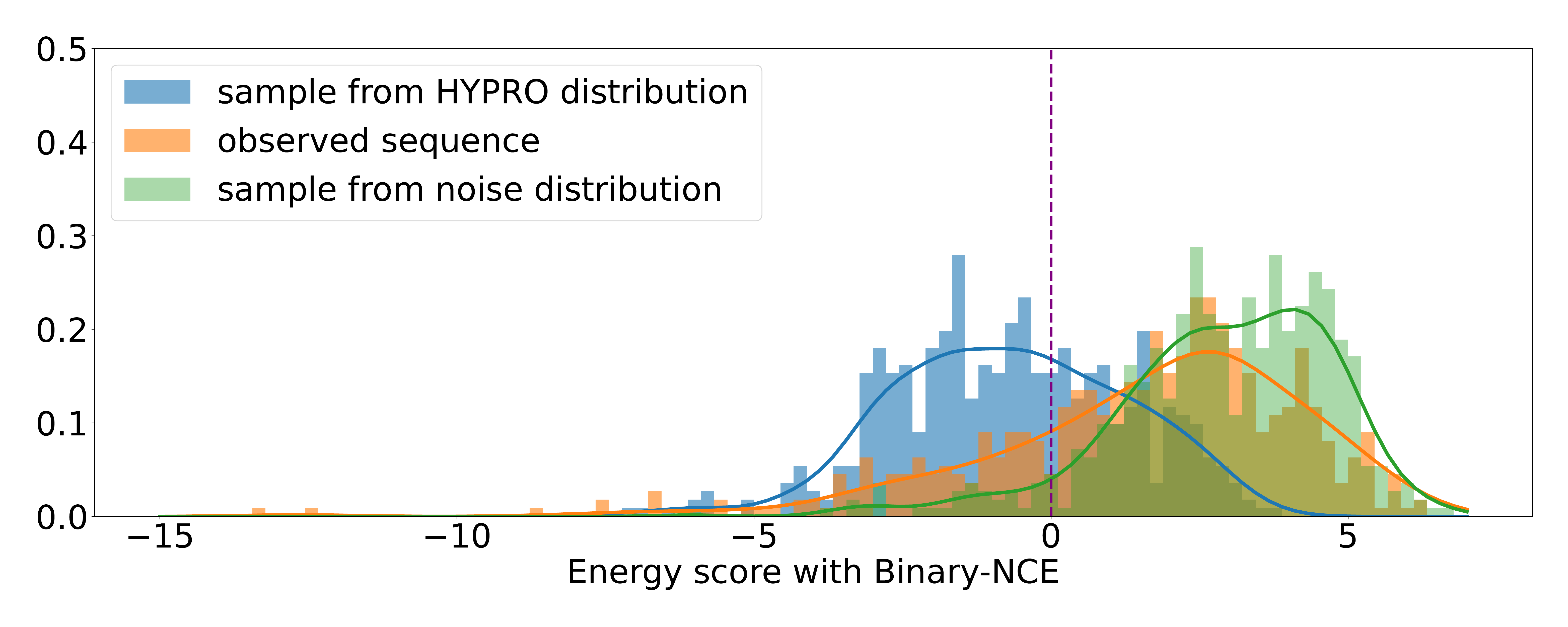}
			~
			\includegraphics[width=0.5\linewidth]{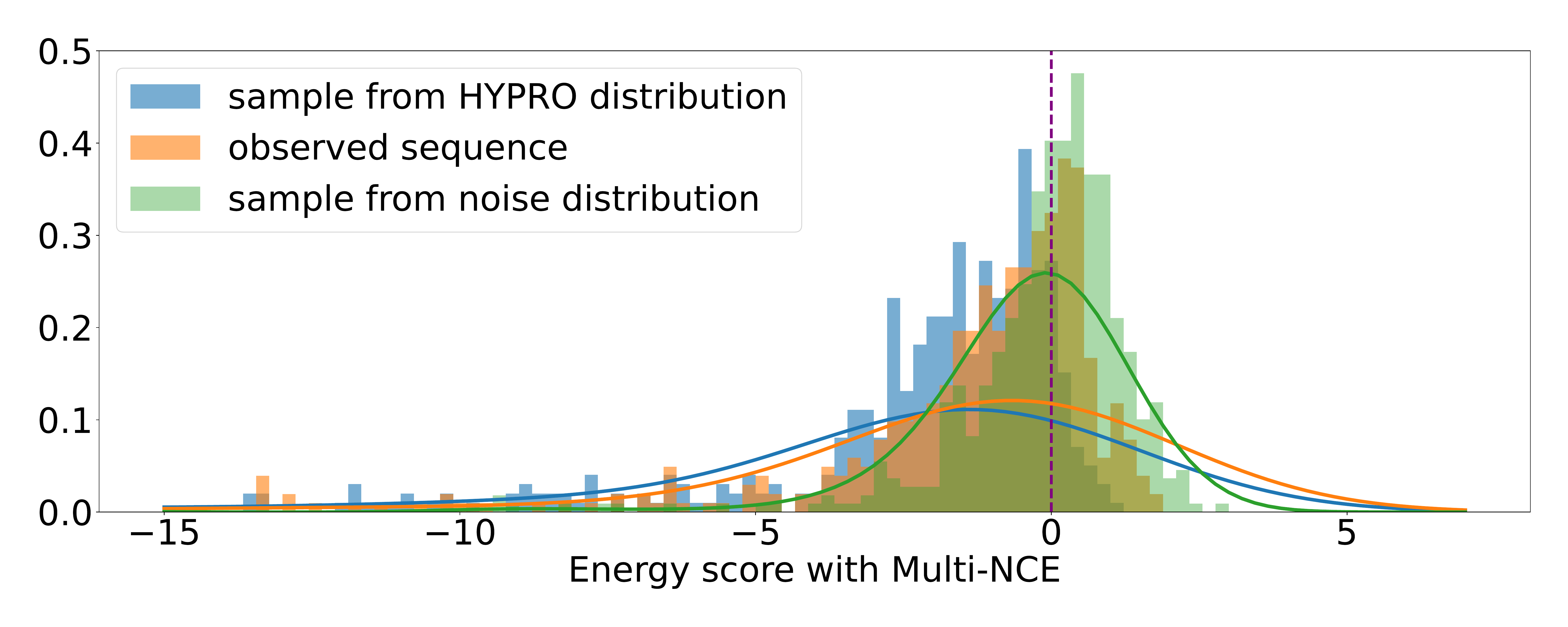}
			\vspace{-20pt}
			\caption{Using the distance-based regularization of \cref{eqn:reg}.}\label{fig:reg_energy_distribution}
		\end{subfigure}
		\vspace{-4pt}
		\caption{Energy scores computed on the held-out development data of Taobao dataset.}\label{fig:energy_distribution}
	\end{center}
\end{figure*}

\paragraph{Analysis-III: Binary-NCE vs.\@ Multi-NCE.}
Both Binary-NCE and Multi-NCE objectives aim to match our full model distribution $p_{\text{HYPRO}}$ with the true data distribution, but Multi-NCE enjoys better statistical properties (see \cref{sec:cont_ebm}) and achieved better performance in our experiments (see \cref{fig:main_results}).
In \cref{fig:unreg_energy_distribution}, we display the distributions of the energy scores of the observed sequences, $p_{\text{HYPRO}}$-generated sequences, and $p_{\text{auto}}$-generated noise sequences: as we can see, the distribution of $p_{\text{HYPRO}}$ is different from that of noise data and indeed closer to that of real data.

\begin{figure*}[t]
	\begin{center}
	    \begin{subfigure}[t]{0.48\linewidth}
            \includegraphics[width=0.47\linewidth]{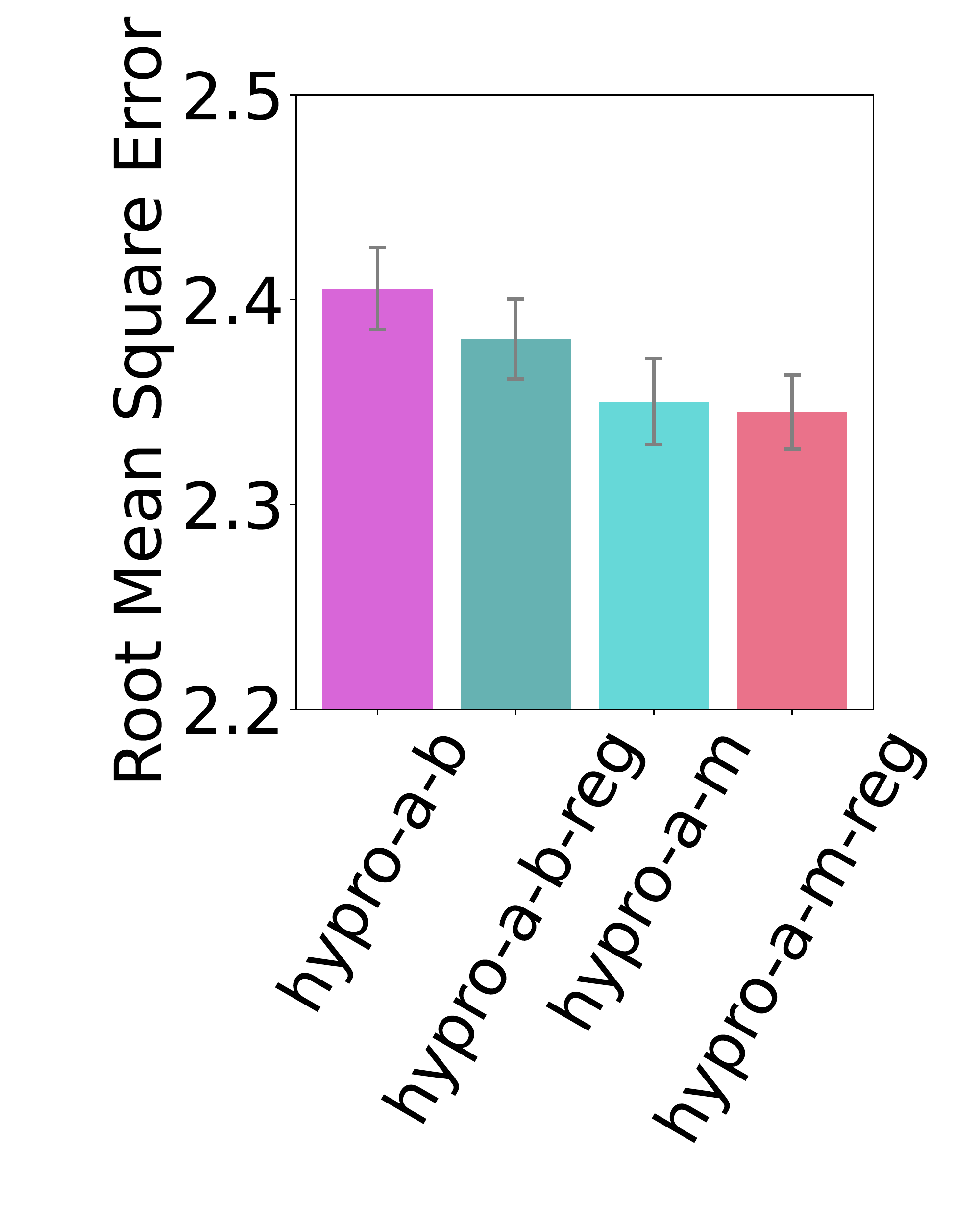}
			~
			\includegraphics[width=0.47\linewidth]{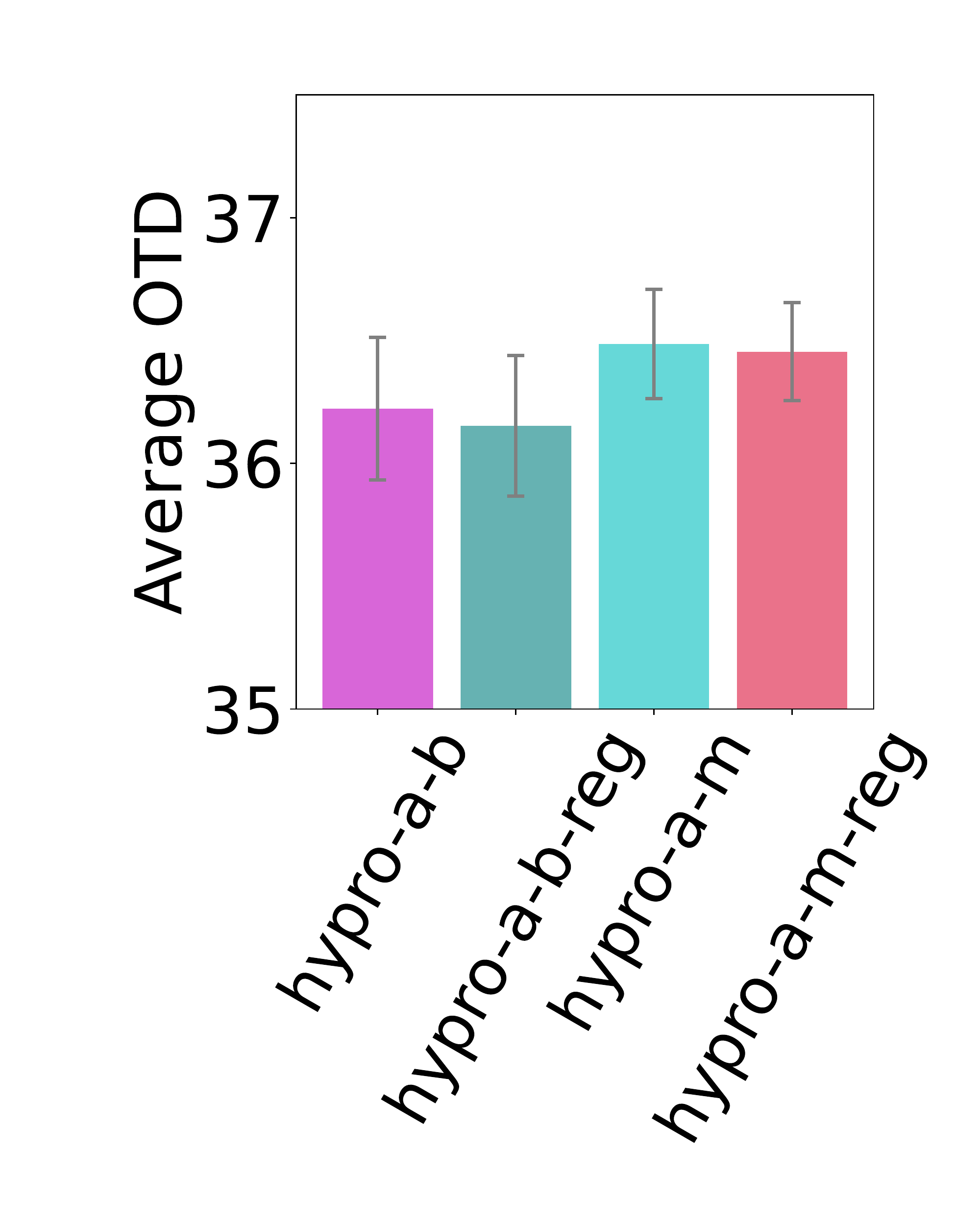}
			\vspace{-10pt}
			\caption{Taobao Dataset.}\label{fig:taobao_reg_effect_compare}
		\end{subfigure}
		~
		\begin{subfigure}[t]{0.48\linewidth}
			\includegraphics[width=0.47\linewidth]{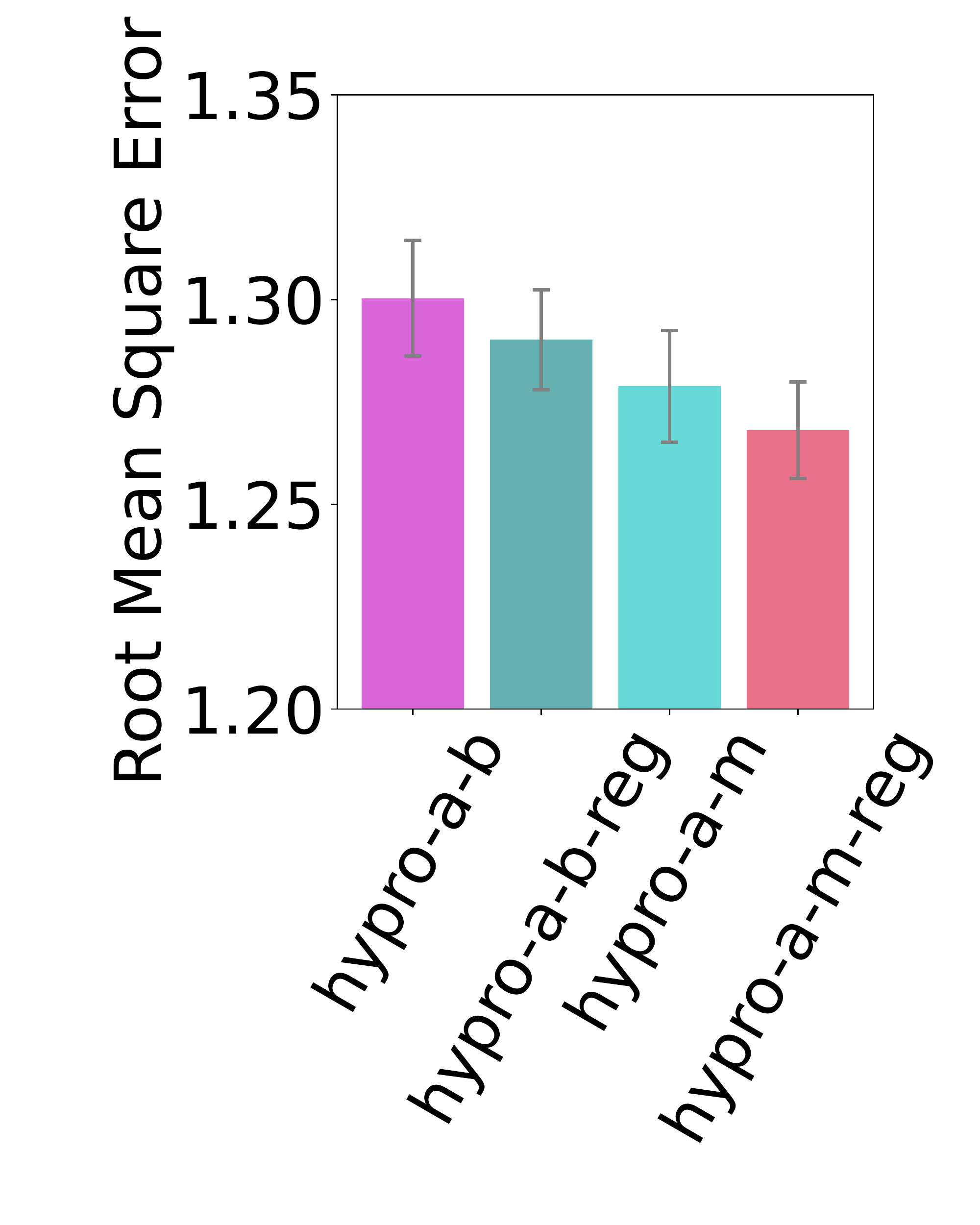}
			~
			\includegraphics[width=0.47\linewidth]{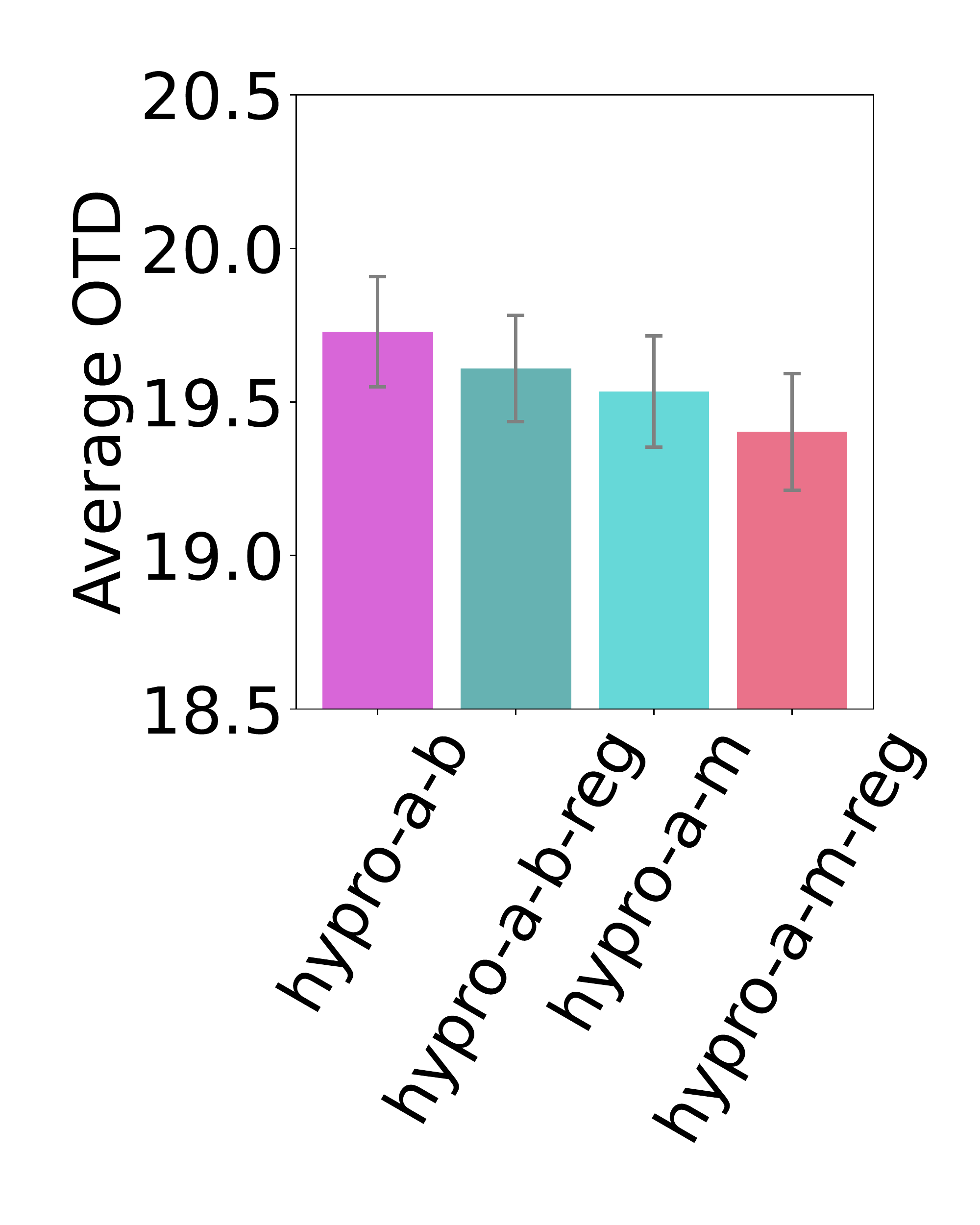}
			\vspace{-10pt}
			\caption{Taxi Dataset.}\label{fig:taxi_reg_effect_compare}
		\end{subfigure}
		\vspace{-4pt}
		\caption{Adding the regularization term $\Omega$. In each figure, the suffix -reg denotes ``with regularization''.}\label{fig:reg_effect_compare}
	\end{center}
 	\vspace{-4pt}
\end{figure*}

\paragraph{Analysis-IV: Effects of the Distance-Based Regularization $\Omega$.}
We experimented with the proposed distance-based regularization $\Omega$ in \cref{eqn:reg}; as shown in \cref{fig:reg_effect_compare}, it slightly improves both Binary-NCE and Multi-NCE. 
As shown in \cref{fig:reg_energy_distribution}, the regularization makes a larger difference in the Binary-NCE case: the energies of $p_{\text{HYPRO}}$-generated sequences are pushed further to the left. 

We did the paired permutation test to verify the statistical significance of our regularization technique; see \cref{app:sigtest} for details. 
Overall, we found that the performance improvements of using the regularization are strongly significant in the Binary-NCE case (p-value $<0.05$ ) but not significant in the Multi-NCE case (p-value $\approx 0.1$). This finding is consistent with the observations in \cref{fig:energy_distribution}.

\begin{figure*}[t]
	\begin{center}
		\includegraphics[width=0.54\linewidth]{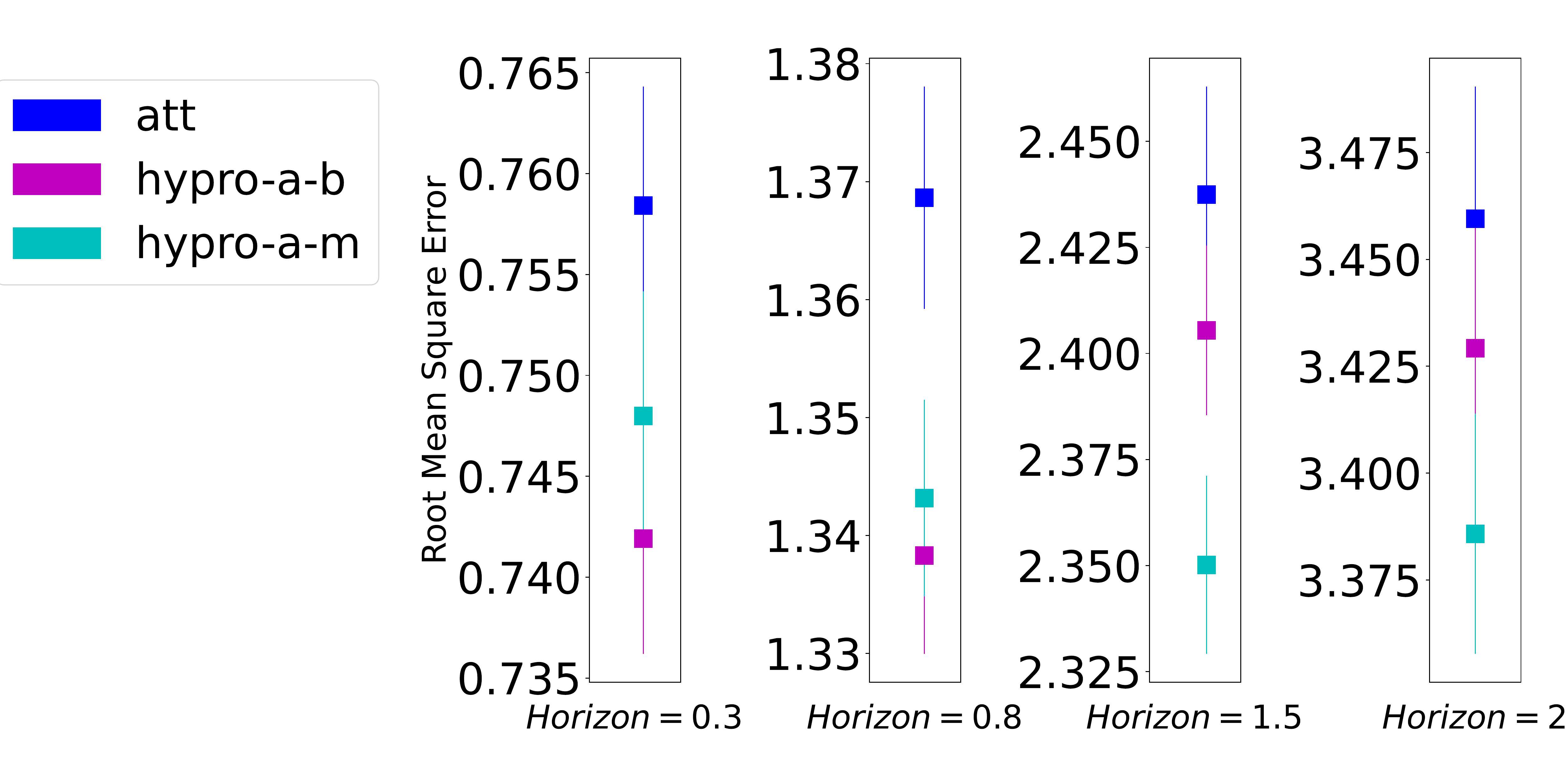}
			~
		\includegraphics[width=0.42\linewidth]{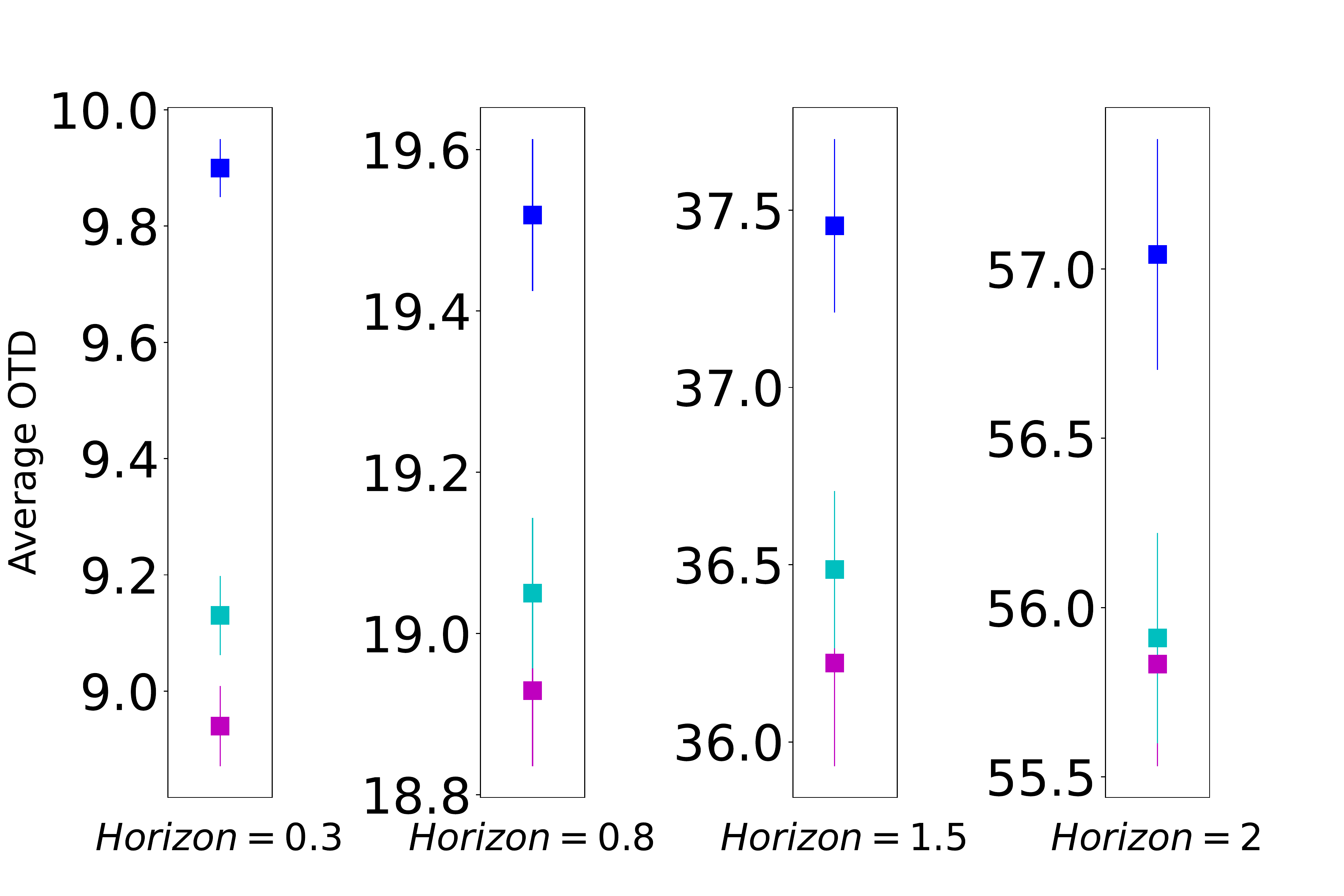}
		\vspace{-4pt}
		\caption{HYPRO vs.\@ AttNHP for different horizons on Taobao Dataset.}\label{fig:results_horizon}
	\end{center}
 	\vspace{-4pt}
\end{figure*}

\paragraph{Analysis-V: Effects of Prediction Horizon.}
\cref{fig:results_horizon} shows how well our method performs for different prediction horizons. On Taobao Dataset, we experimented with horizon being $0.3, 0.8, 1.5, 2$, corresponding to approximately $5, 10, 20, 30$ event tokens, and found that our HYPRO method improves significantly and consistently over the autoregressive baseline AttNHP. 
\begin{figure*}[!htb]
	\begin{center}
	    \begin{subfigure}[t]{0.49\linewidth}
			
            \includegraphics[width=0.47\linewidth]{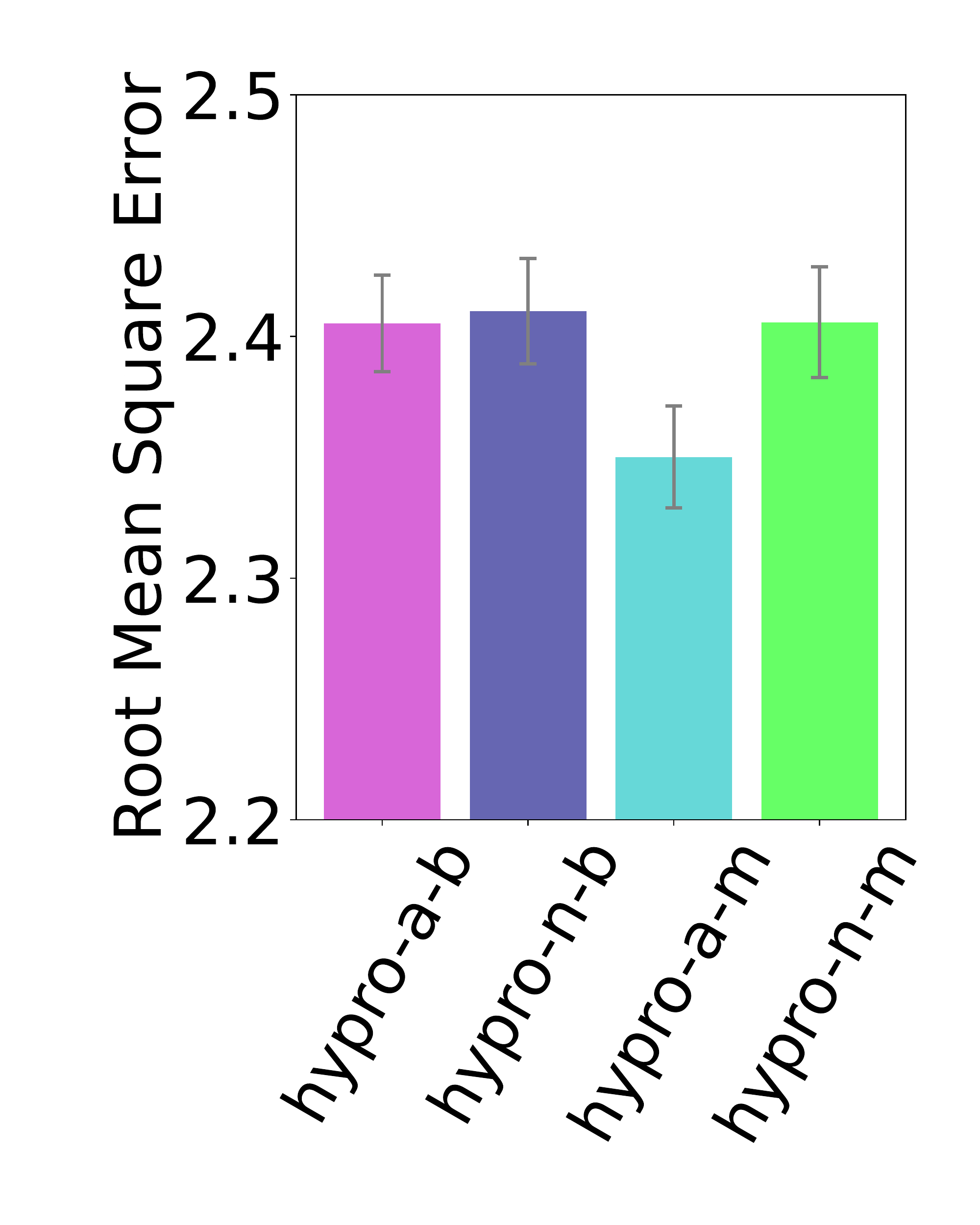}
			~
			\includegraphics[width=0.47\linewidth]{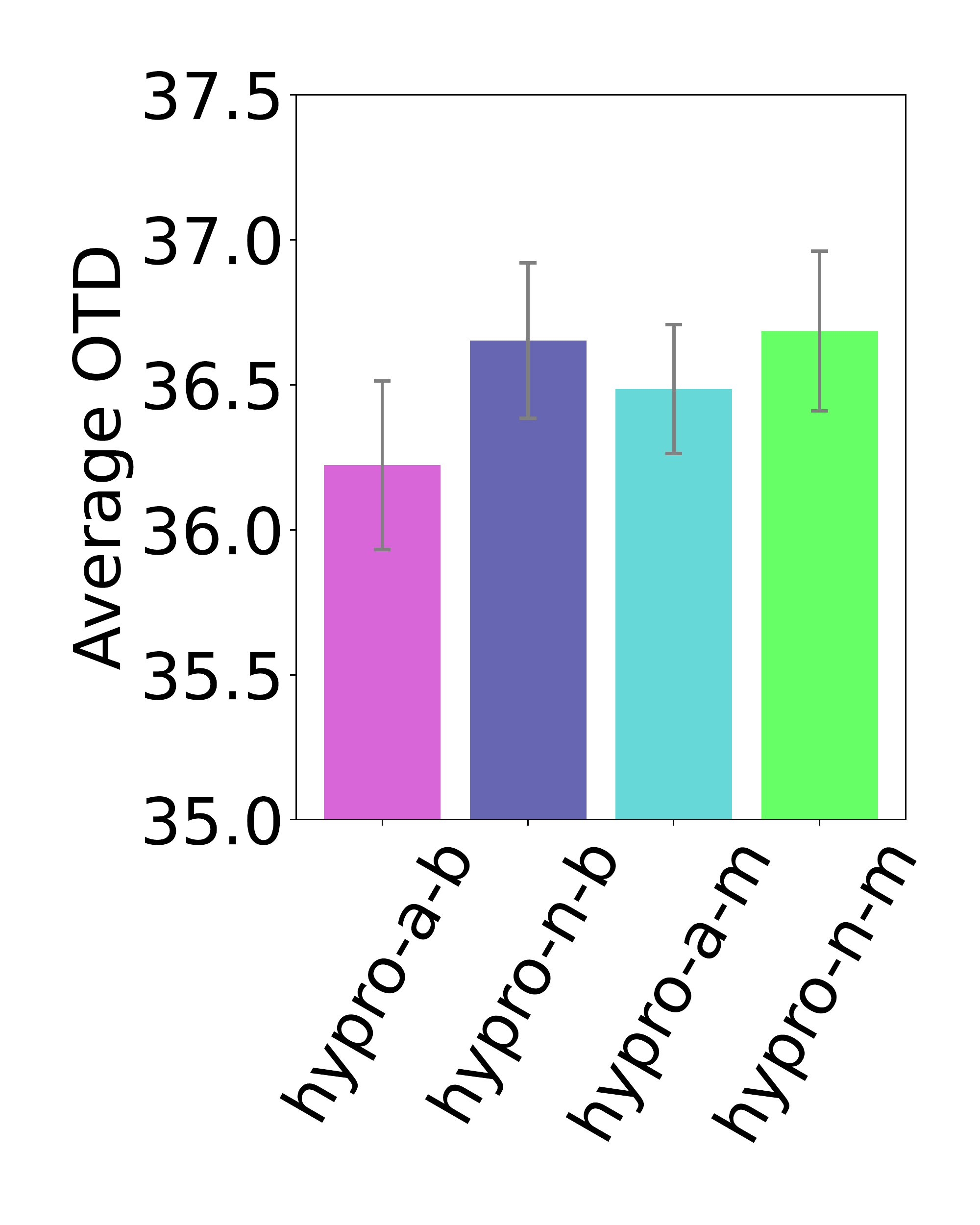}
			
			\vspace{-10pt}
			\caption{Using different energy functions on Taobao Data}\label{fig:modela}
		\end{subfigure}
		~
		\begin{subfigure}[t]{0.49\linewidth}
		
			\includegraphics[width=0.47\linewidth]{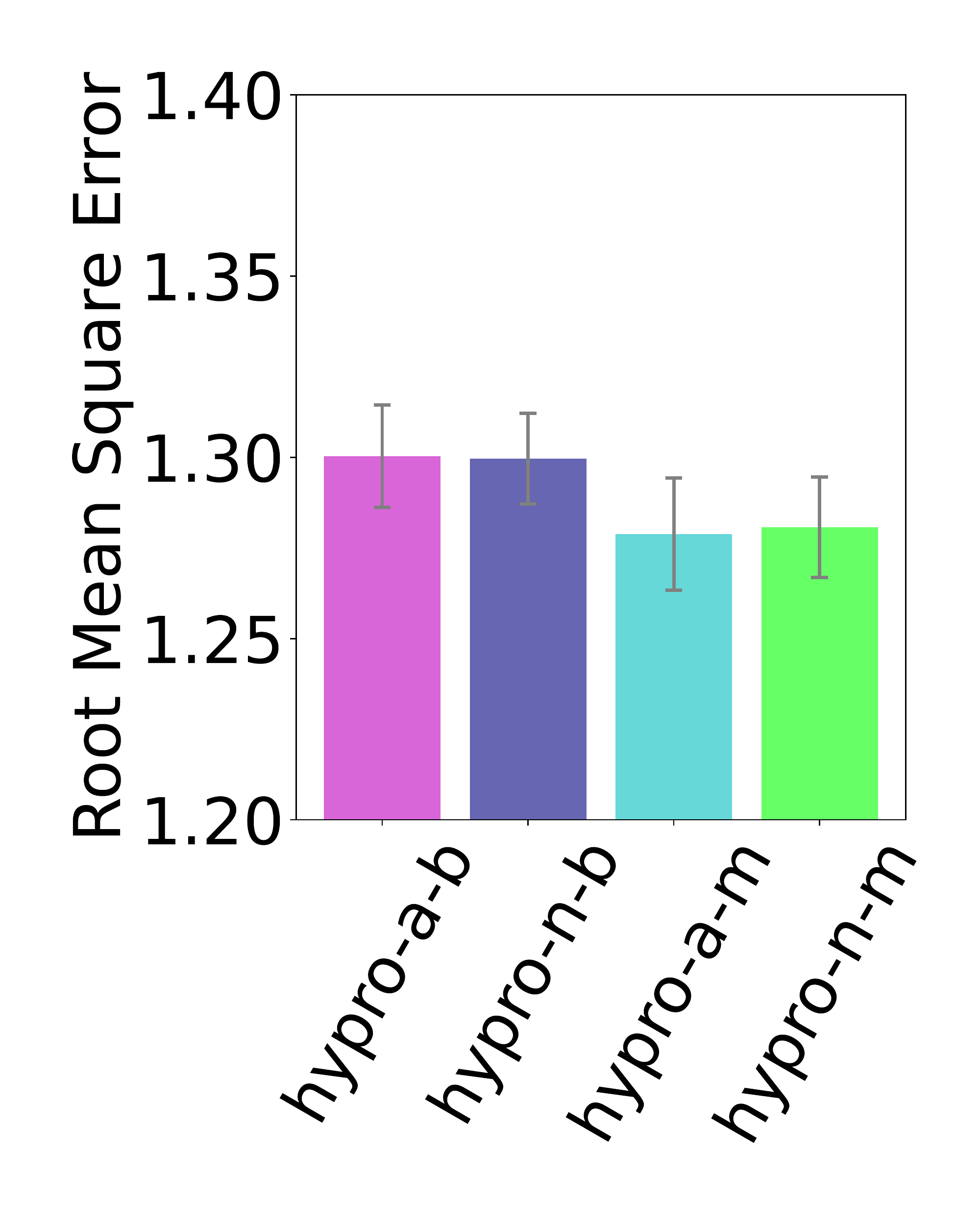}
			~
			\includegraphics[width=0.47\linewidth]{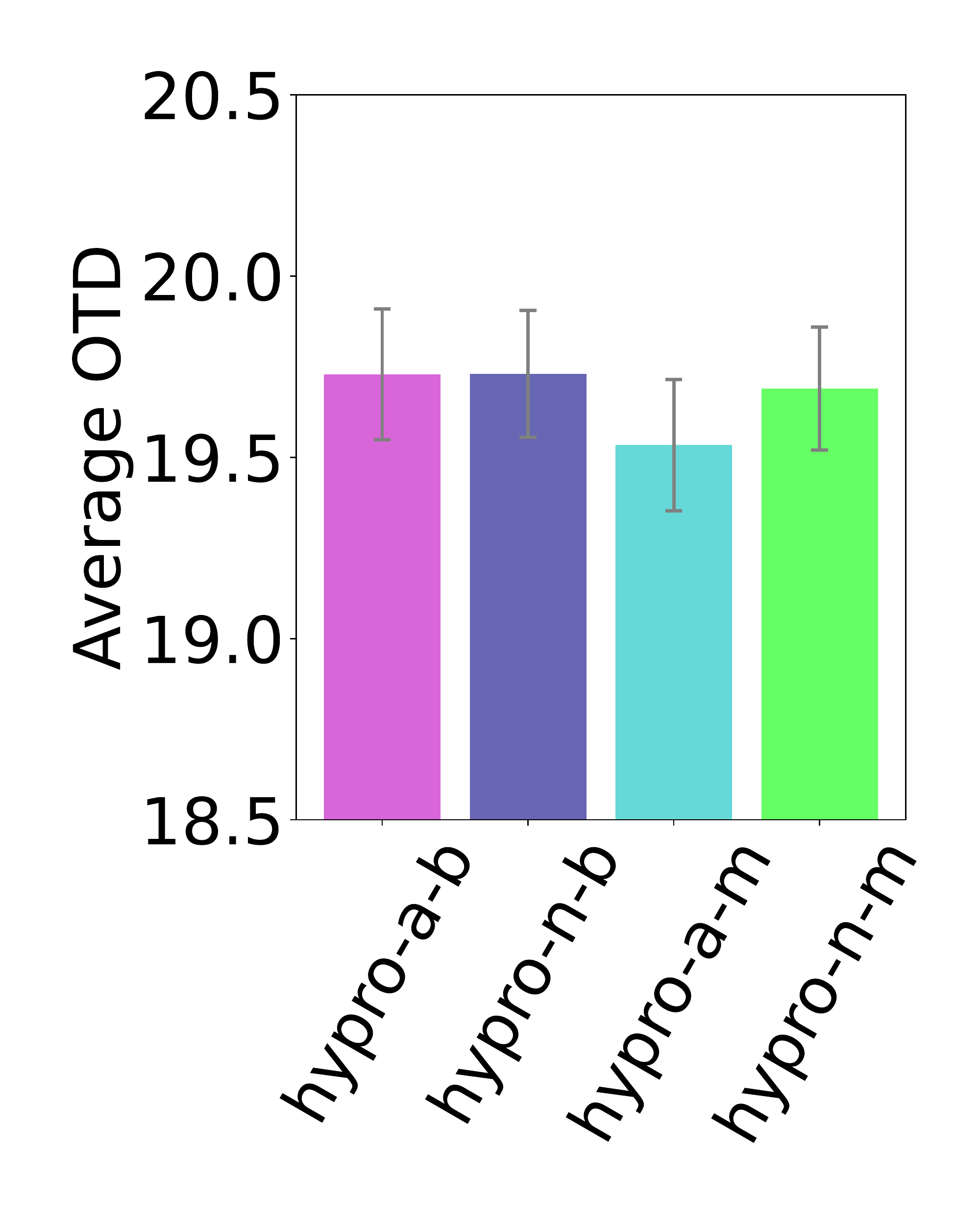}

			\vspace{-10pt}
			\caption{Using different energy functions on Taxi Data}\label{fig:modelb}
		\end{subfigure}
		\vspace{-4pt}
		\caption{Using different energy functions. In each figure, the suffix -n-b and -n-m denote using a continuous-time LSTM as the energy function trained by Binary-NCE and Multi-NCE, respectively.}\label{fig:results_nhp_discriminator}
	\end{center}
% 	\vspace{-12pt}
\end{figure*}

\paragraph{Analysis-VI: Different Energy Functions.} So far we have only shown the results and analysis of using the continuous-time Transformer architecture as the energy function $E_{\param}$. We also experimented with using a continuous-time LSTM~\citep{mei-17-neuralhawkes} as the energy function and found that it never outperformed the Transformer energy function in our experiments; see \cref{fig:results_nhp_discriminator}. We think it is because the Transformer architectures are better at embedding contextual information than LSTMs~\citep{vaswani-2017-transformer,perez2019turing,o2021context}.

\begin{figure*}[!htb]
	\begin{center}
	    \begin{subfigure}[t]{0.49\linewidth}
		
            \includegraphics[width=0.47\linewidth]{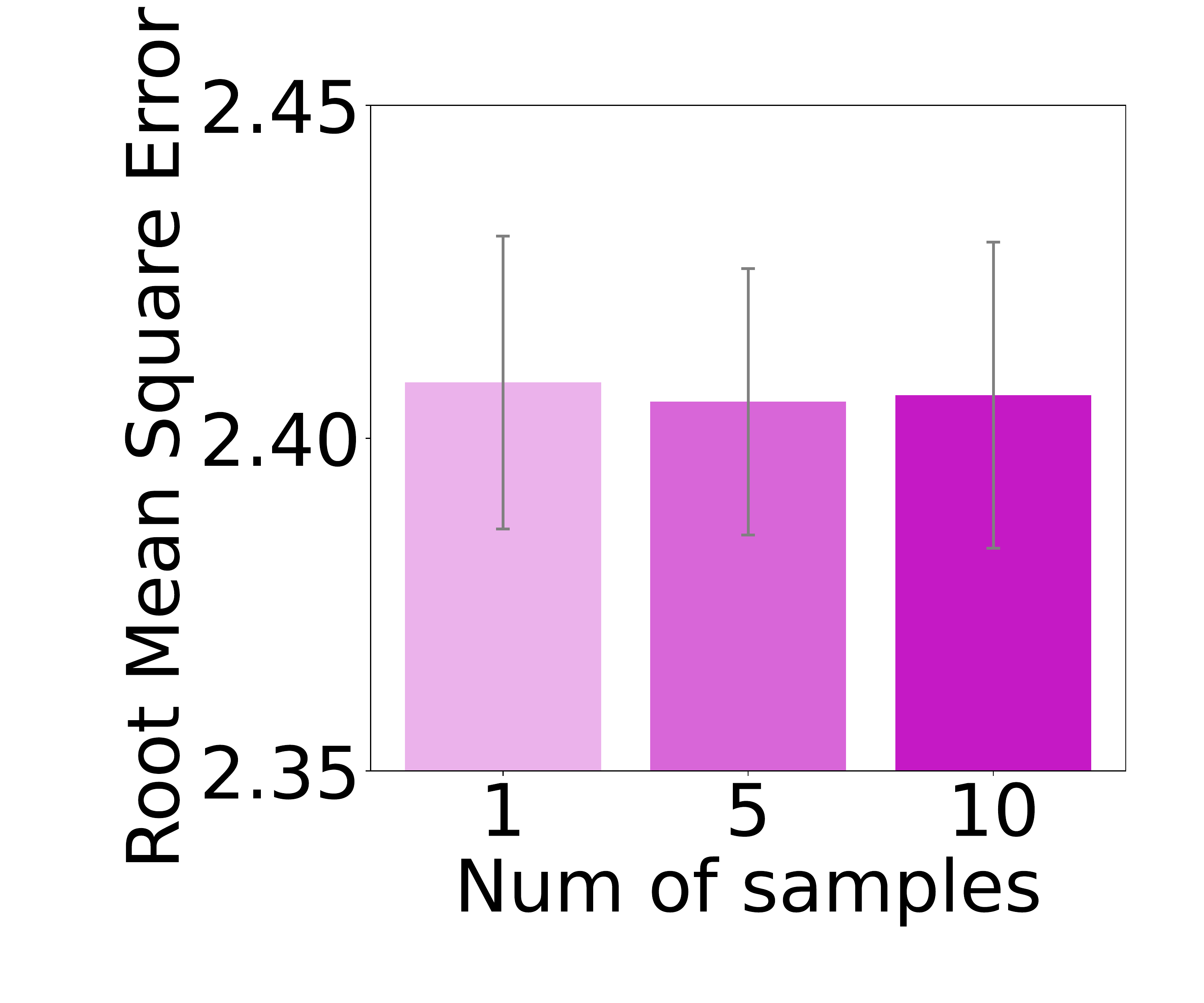}
			~
			\includegraphics[width=0.47\linewidth]{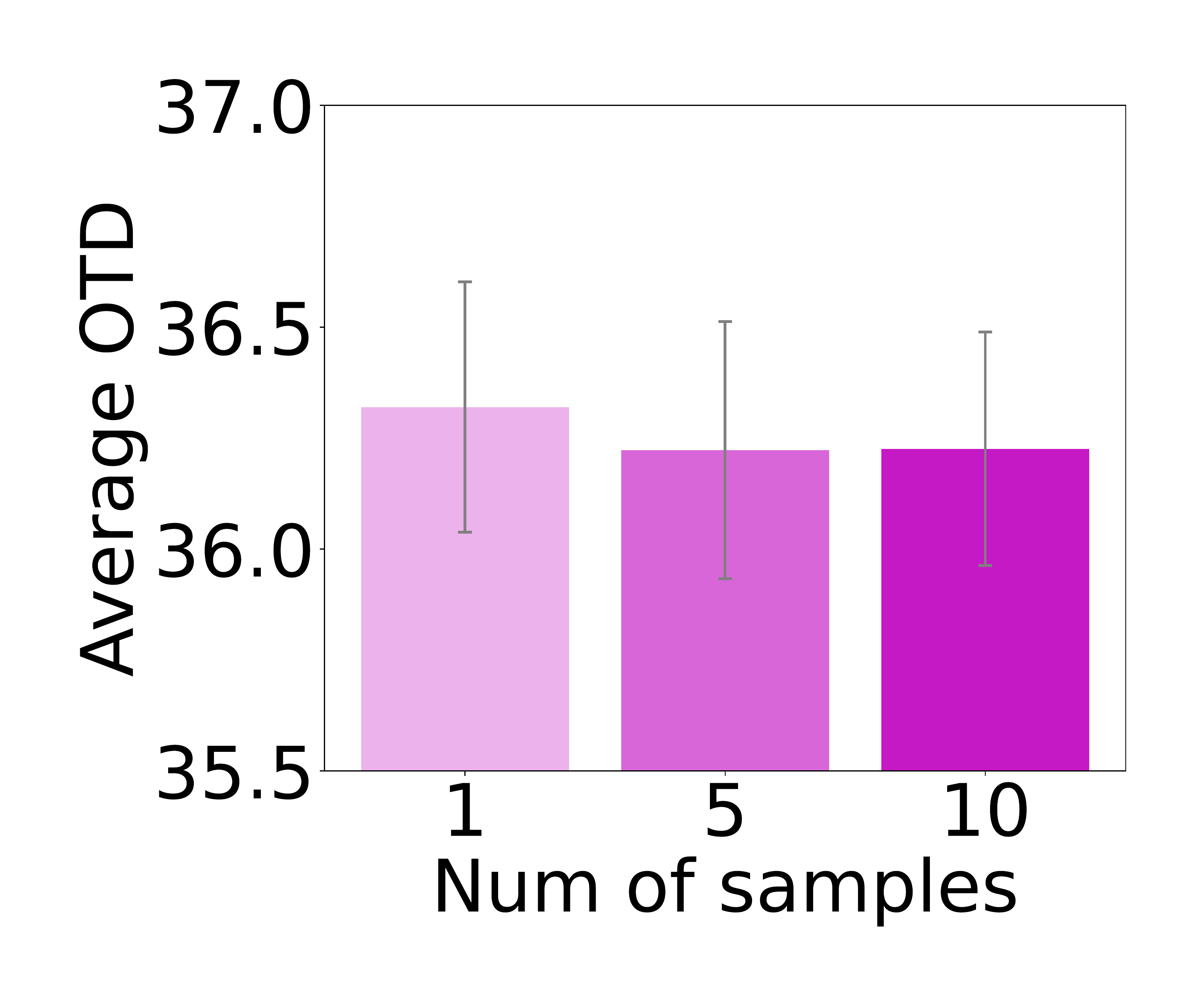}
			
			\vspace{-10pt}
			\caption{Training.}\label{fig:sample_train}
		\end{subfigure}
		~
		\begin{subfigure}[t]{0.49\linewidth}
			
			\includegraphics[width=0.47\linewidth]{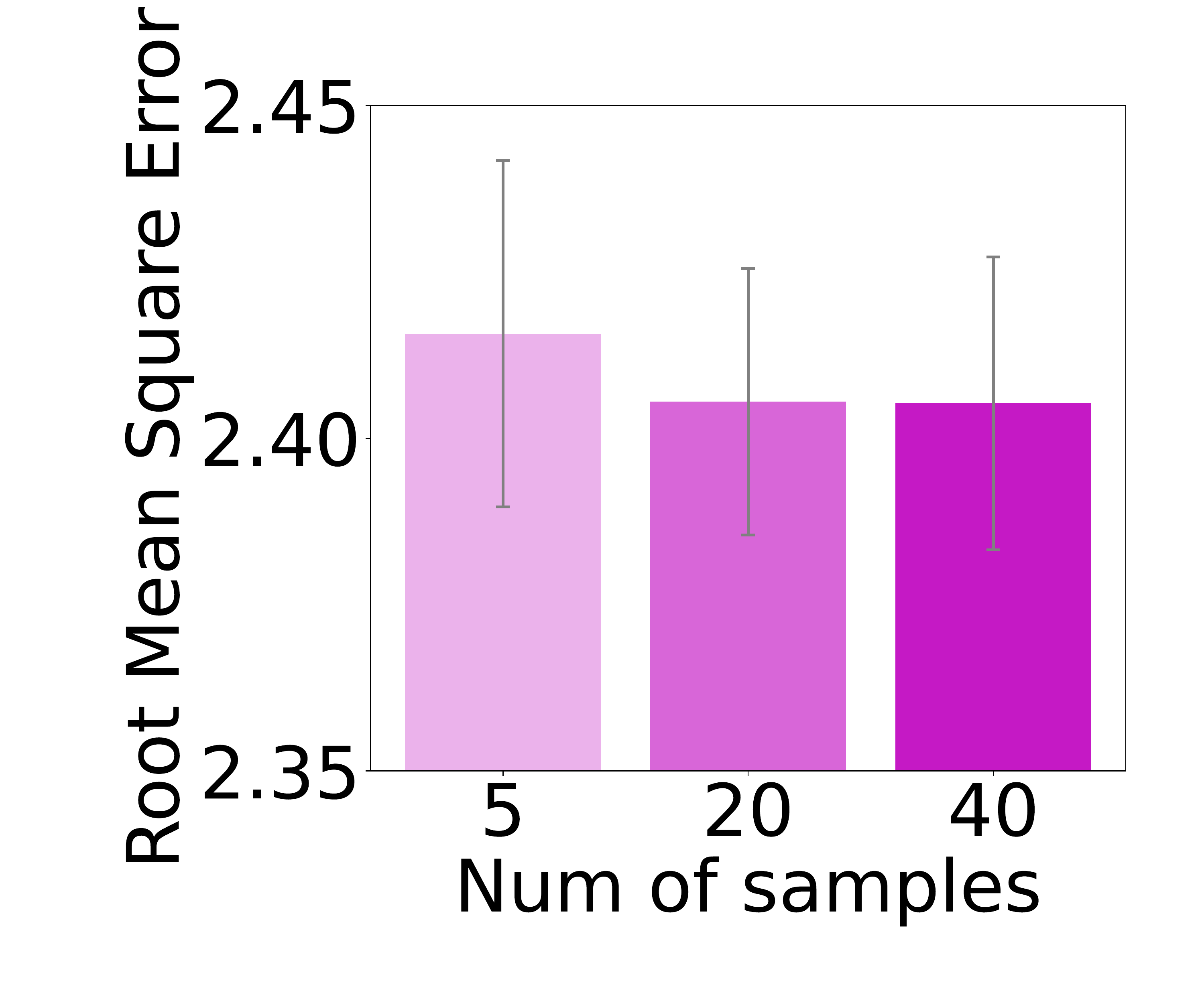}
			~
			\includegraphics[width=0.47\linewidth]{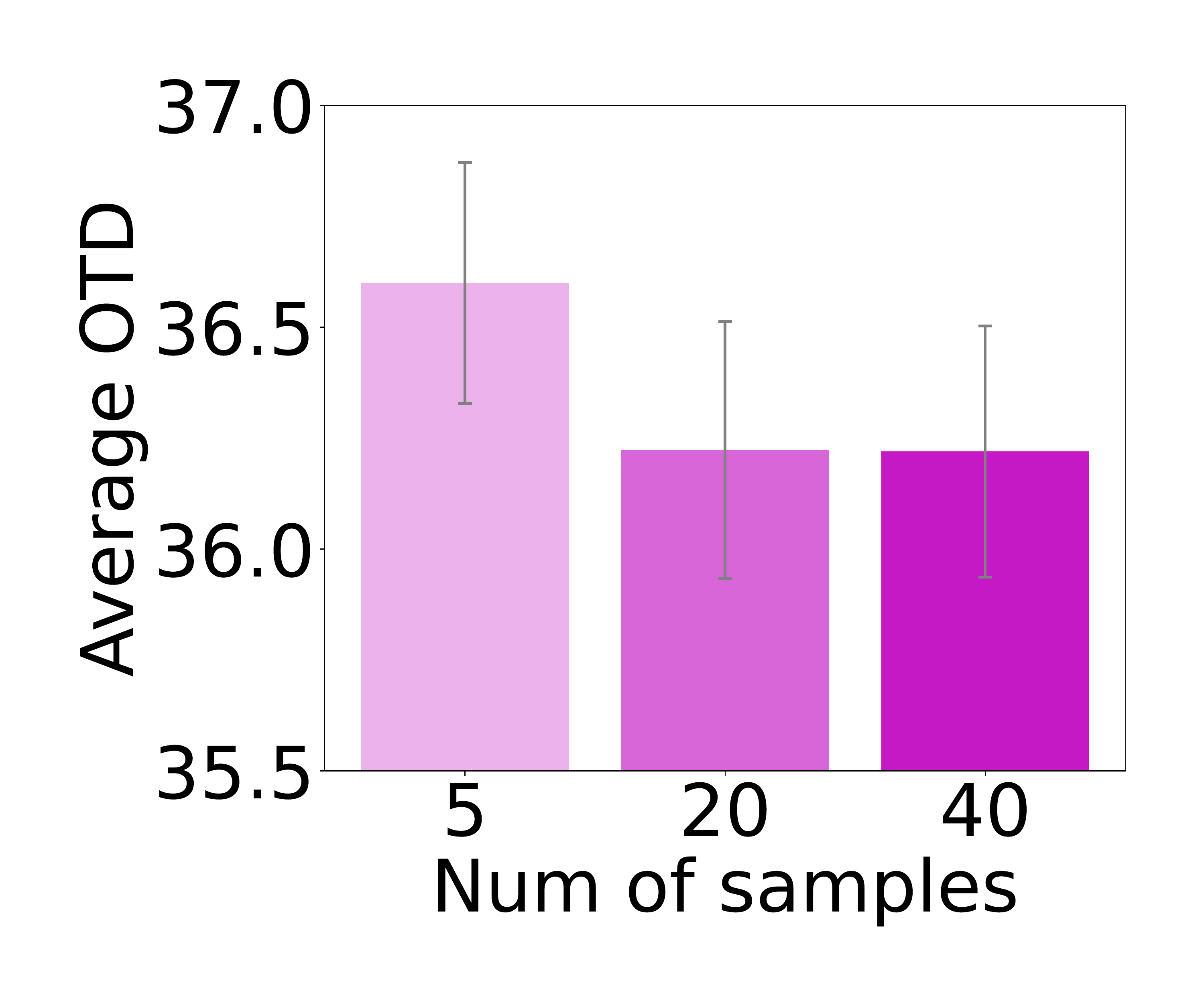}
			\vspace{-10pt}
			\caption{Inference.}\label{fig:sample_test}
		\end{subfigure}
		\vspace{-4pt}
		\caption{Effects of the number of negative samples in training and inference on the Taobao dataset.}\label{fig:results_num_samples}
	\end{center}
\end{figure*}

\paragraph{Analysis-VII: Negative Samples.} 
On the Taobao dataset, we analyzed how the number of negative samples affects training and inference. We experimented with the Binary-NCE objective without the distance regularization (i.e., hypro-a-b). 
The results are in \cref{fig:results_num_samples}. 
During training, increasing the number of negative samples from 1 to 5 has brought improvements but further increasing it to 10 does not. 
During inference, the results are improved when we increase the number of samples from 5 to 20, but they stop improving when we further increase it. Throughout the paper, we used 5 in training and 20 in inference (\cref{app:training_details}).

\section{Conclusion}
We presented HYPRO, a hybridly normalized neural probabilistic model for the task of long-horizon prediction of event sequences. 
Our model consists of an autoregressive base model and an energy function: the latter learns to reweight the sequences drawn from the former such that the sequences that appear more realistic as a whole can end up with higher probabilities under our full model. 
We developed two training objectives that can train our model without computing its normalizing constant, together with an efficient inference algorithm based on normalized importance sampling. 
Empirically, our method outperformed current state-of-the-art autoregressive models as well as a recent non-autoregressive model designed specifically for the same task. 

\section{Limitations and Societal Impacts}\label{sec:limit}\label{sec:impact}
\paragraph{Limitations.} Our method uses neural networks, which are typically data-hungry. Although it worked well in our experiments, it might still suffer compared to non-neural models if starved of data. 
Additionally, our method requires the training sequences to be sufficiently long so it can learn to make long-horizon predictions; it may suffer if the training sequences are short.

\paragraph{Societal Impacts.} Our paper develops a novel probabilistic model for long-horizon prediction of event sequences. By describing the model and releasing code, we hope to facilitate probabilistic modeling of continuous-time sequential data in many domains. 
However, like many other machine learning models, our model may be applied to unethical ends. For example, its abilities of better fitting data and making more accurate predictions could potentially be used for unwanted tracking of individual behavior, e.g. for surveillance.

\section*{Acknowledgments}
This work was supported by a research gift to the last author by Adobe Research. 
We thank the anonymous NeurIPS reviewers and meta-reviewer for their constructive feedback. 
We also thank our colleagues at Ant Group for helpful discussion.

\clearpage
\bibliographystyle{icml2020_url}
\vspace{0pt}
\bibliography{reb_tpp}

\clearpage
\section*{Checklist}

\begin{enumerate}

\item For all authors...
    \begin{enumerate}
      \item Do the main claims made in the abstract and introduction accurately reflect the paper's contributions and scope?
        \answerYes{}
      \item Did you describe the limitations of your work?
        \answerYes{See \cref{sec:limit}}
      \item Did you discuss any potential negative societal impacts of your work?
        \answerYes{See \cref{sec:impact}.}
      \item Have you read the ethics review guidelines and ensured that your paper conforms to them?
        \answerYes{}
    \end{enumerate}

\item If you are including theoretical results...
\begin{enumerate}
  \item Did you state the full set of assumptions of all theoretical results?
    \answerYes{See \cref{sec:prilim,sec:cont_ebm,app:nce}.}
        \item Did you include complete proofs of all theoretical results?
    \answerYes{See \cref{app:nce}.}
\end{enumerate}

\item If you ran experiments...
\begin{enumerate}
  \item Did you include the code, data, and instructions needed to reproduce the main experimental results (either in the supplemental material or as a URL)?
    \answerYes{See the supplemental material.}
  \item Did you specify all the training details (e.g., data splits, hyperparameters, how they were chosen)?
    \answerYes{See \cref{app:data_stat} and \cref{app:training_details}.}
        \item Did you report error bars (e.g., with respect to the random seed after running experiments multiple times)?
    \answerYes{See \cref{sec:exp}.}
        \item Did you include the total amount of compute and the type of resources used (e.g., type of GPUs, internal cluster, or cloud provider)?
    \answerYes{See \cref{app:training_details}.}
\end{enumerate}

\item If you are using existing assets (e.g., code, data, models) or curating/releasing new assets...
\begin{enumerate}
  \item If your work uses existing assets, did you cite the creators?
    \answerYes{See \cref{app:training_details}}
  \item Did you mention the license of the assets?
    \answerYes{See \cref{app:imp}.}
  \item Did you include any new assets either in the supplemental material or as a URL?
    \answerYes{We include our code in the supplementary material.} 
  \item Did you discuss whether and how consent was obtained from people whose data you're using/curating?
    \answerNA{We use existing datasets that are publicly available and already anonymized.}
  \item Did you discuss whether the data you are using/curating contains personally identifiable information or offensive content?
    \answerNA{We use existing datasets that are publicly available and already anonymized.}
\end{enumerate}

\item If you used crowdsourcing or conducted research with human subjects...
\begin{enumerate}
  \item Did you include the full text of instructions given to participants and screenshots, if applicable?
    \answerNA{}
  \item Did you describe any potential participant risks, with links to Institutional Review Board (IRB) approvals, if applicable?
    \answerNA{}
  \item Did you include the estimated hourly wage paid to participants and the total amount spent on participant compensation?
    \answerNA{}
\end{enumerate}

\end{enumerate}

%%%%%%%%%%%%%%%%%%%%%%%%%%%%%%%%%%%%%%%%%%%%%%%%%%%%%%%%%%%%

\clearpage
\appendix
\appendixpage

\section{Method Details}\label{app:method}

\subsection{Derivation of NCE Objectives}\label{app:nce}

Given a prefix $\es{x}{[0,T]}$, we have the true continuation $\es{x}{(T,T']}^{(0)}$ and $N$ noise samples $\es{x}{(T,T']}^{(1)}, \ldots, \es{x}{(T,T']}^{(N)}$. 
By concatenating the prefix and each (true or noise) continuation, we obtain $N+1$ completed sequences $\es{x}{[0,T']}^{(0)}, \es{x}{[0,T']}^{(1)}, \ldots, \es{x}{[0,T']}^{(N)}$. 

\paragraph{Binary-NCE Objective.}
For each completed sequence $\es{x}{[0,T']}^{(n)}$, we learn to classify whether it is real data or noise data. 
The unnormalized probability for each case is: 
\begin{align}
    \tilde{p}\left( \text{it is real data} \right) 
    &= p_{\text{HYPRO}}\left(\es{x}{(T,T']}^{(n)} \mid \es{x}{[0,T]} \right) = p_{\text{auto}}\left( \es{x}{(T, T']}^{(n)} \mid \es{x}{[0, T]} \right) \frac{\exp\left(- E_{\param}(\es{x}{[0, T']}^{(n)}) \right)}{Z_{\param} \left( \es{x}{[0, T]} \right)} \\
    \tilde{p}\left( \text{it is noise data} \right) 
    &= p_{\text{auto}}\left( \es{x}{(T, T']}^{(n)} \mid \es{x}{[0, T]} \right)
    %p_{\text{HYPRO}}\left( \es{x}{(T, T']} \mid \es{x}{[0, T)} \right) = p_{\text{auto}}\left( \es{x}{(T, T']} \mid \es{x}{[0, T]} \right) \frac{\exp\left(- E_{\param}(\es{x}{[0, T']}) \right)}{Z_{\param} \left( \es{x}{[0, T]} \right)}
\end{align}
Then the normalized probabilities are: 
\begin{align}
    p\left( \text{it is real data} \right) 
    &= \frac{\tilde{p}\left( \text{it is real data} \right) }{\tilde{p}\left( \text{it is real data} \right)  + \tilde{p}\left( \text{it is noise data} \right) }
    = \frac{\exp\left(- E_{\param}(\es{x}{[0, T']}^{(n)}) \right)}{Z_{\param} \left( \es{x}{[0, T]} \right) + \exp\left(- E_{\param}(\es{x}{[0, T']}^{(n)}) \right)} \\
    p\left( \text{it is noise data} \right) 
    &= \frac{\tilde{p}\left( \text{it is noise data} \right) }{\tilde{p}\left( \text{it is real data} \right)  + \tilde{p}\left( \text{it is noise data} \right) }
    = \frac{Z_{\param} \left( \es{x}{[0, T]} \right)}{Z_{\param} \left( \es{x}{[0, T]} \right) + \exp\left(- E_{\param}(\es{x}{[0, T']}^{(n)}) \right)}
\end{align}
Following previous work~\citep{mnih-12-nce}, we assume that the model is self-normalized, i.e., $Z_{\param} \left( \es{x}{[0, T]} \right) = 1$. 
Then the normalized probabilities become
\begin{align}
    p\left( \text{it is real data} \right) 
    &= \frac{\exp\left(- E_{\param}(\es{x}{[0, T']}^{(n)}) \right)}{1 + \exp\left(- E_{\param}(\es{x}{[0, T']}^{(n)}) \right)} 
    = \sigma \left( -E_{\param}(\es{x}{[0, T']}^{(n)}) \right) \\
    p\left( \text{it is noise data} \right) 
    &= \frac{1}{1 + \exp\left(- E_{\param}(\es{x}{[0, T']}^{(n)}) \right)}
    = \sigma \left( E_{\param}(\es{x}{[0, T']}^{(n)}))\right) 
    %J_{\text{binary}} &= \log \sigma \left( -E_{\param}(\es{x}{[0, T']}^{(0)}) \right) + \sum_{n=1}^{N} \log \sigma \left( E_{\param}(\es{x}{[0, T']}^{(n)}))\right)
\end{align}
where $\sigma$ is the sigmoid function. 

For the true completed sequence $\es{x}{[0,T']}^{(0)}$, we maximize the log probability that it is real data, i.e., $\log p\left( \text{it is real data} \right)$; for each noise sequence $\es{x}{[0,T']}^{(n)}$, we maximize the log probability that it is noise data, i.e., $\log p\left( \text{it is noise data} \right)$. The Binary-NCE objective turns out to be \cref{eqn:bin_entropy_loss}, i.e., 
\begin{align*}
    J_{\text{binary}} &= \log \sigma \left( -E_{\param}(\es{x}{[0, T']}^{(0)}) \right) + \sum_{n=1}^{N} \log \sigma \left( E_{\param}(\es{x}{[0, T']}^{(n)}))\right)
\end{align*}

\paragraph{Multi-NCE Objective.}
For these $N+1$ sequences, we learn to discriminate the true sequence against the noise sequences. 
For each of them $\es{x}{[0,T']}^{(n)}$, the following is the unnormalized probability that it is real data but all others are noise: 
\begin{align}
    \tilde{p}\left( \es{x}{[0,T']}^{(n)} \text{ is real, others are noise} \right) 
    &= p_{\text{HYPRO}}\left(\es{x}{(T,T']}^{(n)} \mid \es{x}{[0,T]} \right) \prod_{n'\neq n} p_{\text{auto}}\left( \es{x}{(T, T']}^{(n')} \mid \es{x}{[0, T]} \right)
\end{align}
where can be rearranged to be 
\begin{align}
    \tilde{p}\left( \es{x}{[0,T']}^{(n)} \text{ is real, others are noise} \right) 
    &= \frac{\exp\left(- E_{\param}(\es{x}{[0, T']}^{(n)}) \right)}{Z_{\param} \left( \es{x}{[0, T]} \right)} \prod_{n=0}^{N} p_{\text{auto}}\left( \es{x}{(T, T']}^{(n)} \mid \es{x}{[0, T]} \right)
\end{align}
Note that $\frac{1}{Z}\prod_{n=0}^{N} p_{\text{auto}}$ is constant with respect to $n$. So we can ignore that term and obtain
\begin{align}
    \tilde{p}\left( \es{x}{[0,T']}^{(n)} \text{ is real, others are noise} \right) 
    &\propto \exp\left(- E_{\param}(\es{x}{[0, T']}^{(n)}) \right)
\end{align}
Therefore, we can obtain the normalized probability that $\es{x}{[0,T']}^{(0)}$ is real data as below
\begin{align}
    p\left( \es{x}{[0,T']}^{(0)} \text{ is real data} \right) 
    &= \frac{\exp\left( - E_{\param}(\es{x}{[0, T']}^{(0)}))\right)}{\sum_{n=0}^{N} \exp\left( - E_{\param}(\es{x}{[0, T']}^{(n)}))\right)} 
\end{align}
Note that the normalizing constant $Z$ doesn't show up in the normalized probability since it has been cancelled out as a part of the $\frac{1}{Z}\prod_{n=0}^{N} p_{\text{auto}}$ constant. 
That is, unlike the Binary-NCE case, we do not need to assume self-normalization in this Multi-NCE case. 

We maximize the log probability that $\es{x}{[0,T']}^{(0)}$ is real data, i.e., $\log p\left( \es{x}{[0,T']}^{(0)} \text{ is real data} \right)$; the \mbox{Multi-NCE} objective turns out to be \cref{eqn:multi_entropy_loss}, i.e., 
\begin{align*}
    J_{\text{multi}} &=  -E_{\param}(\es{x}{[0, T']}^{(0)}) - \log \sum_{n=0}^{N} \exp\left( - E_{\param}(\es{x}{[0, T']}^{(n)}))\right)
\end{align*}

\subsection{Sampling Algorithm Details}\label{app:thinning}
In \cref{sec:infer}, we described a sampling method to approximately draw $\es{x}{(T,T']}$ from $p_{\text{HYPRO}}$. 
It calls the thinning algorithm, which we describe in \cref{alg:thinning}. 
\begin{algorithm}[tb]
    \caption{Thinning Algorithm.}\label{alg:thinning}
    \begin{algorithmic}[1]
		\INPUT an event sequence $\es{x}{[0,T]}$ over the given interval $[0,T]$ and an interval $(T, T']$ of interest; \newline
		trained autoregressive model $p_{\text{auto}}$ 
		\OUTPUT a sampled continuation $\es{x}{(T,T']}$
		\Procedure{Thinning}{$\es{x}{[0,T]}, T', p_{\text{auto}}$}
		\State initialize $\es{x}{(T,T']}$ as empty
		\LineComment{use the thinning algorithm to draw each noise sequences from the autoregressive model $p_{\text{auto}}$}
	    \State $t_0\gets T$; $i\gets 1$; $\history \gets \es{x}{[0,T]}$
	
		\While{$t_0 < T'$} \Comment{draw next event if we haven't exceeded the time boundary $T'$ yet}
		\LineComment{upper bound $\lambda^*$ can be found for NHP and AttNHP.}
		\LineComment{technical details can be found in \citet{mei-17-neuralhawkes} and \citet{yang-2022-transformer}.}
		\State find upper bound $\lambda^* \geq \sum_{k=1}^{K}\lambda_k(t \mid \history )$ for all $t \in (t_{0},\infty)$ \Comment{compute sampling intensity}
		\Repeat
	    \State draw $\Delta \sim \Exp(\lambda^*)$; $t_0 \pluseq \Delta$ \Comment{time of next proposed noise event}
			\State $u \sim \Uniform(0,1)$ 
		\Until{$u\lambda^* \leq \sum_{k=1}^{K}\lambda_k(t_0\mid\history)$ } \Comment{accept proposed next noise event with prob $\sum_{k=1}^{K}\lambda_{k}/\lambda^*$}
		\IfThen{$t_0 > T'$}{{\bfseries break}}
		\State draw $k \in \{1, \ldots, K\}$ where probability of $k$ is $\propto \lambda_k(t_0 \mid \history)$
		\State append $(t_0, k)$ to both $\history$ and $\es{x}{(T,T']}$
		\EndWhile
		\State \textbf{return} $\es{x}{(T,T']}$
		\EndProcedure
	\end{algorithmic}
\end{algorithm}

\section{Experimental Details}

\subsection{Dataset Details}\label{app:data_stat}

{\bfseries Taobao} \citep{tianchi-taobao-2018}. This dataset contains time-stamped user click behaviors on Taobao shopping pages from November 25 to December 03, 2017. Each user has a sequence of item click events with each event containing the timestamp and the category of the item. The categories of all items are first ranked by frequencies and the top $16$ are kept while the rests are merged into one category, with each category corresponding to an event type. We work on a subset of $2000$ most active users with average sequence length $58$ and then end up with $K=17$ event types. We randomly sampled disjoint train, dev and test sets with $1300$, $200$ and $500$ sequences from the dataset. Given the average inter-arrival time $0.06$ (time unit is $3$ hours), we choose the prediction horizon as $1.5$ that approximately has $20$ event tokens per sequence.

{\bfseries Taxi} \citep{whong-14-taxi}. This dataset contains time-stamped taxi pickup and drop off events with zone location ids in New York city in 2013 .
Following the processing recipe of previous work \citep{mei-19-smoothing},  each event type is defined as a tuple of (location, action).  The location
is one of the 5 boroughs $\{$Manhattan,
Brooklyn, Queens, The Bronx, Staten Island$\}$. The action can be either pick-up or drop-off. Thus, there are $K=5\times 2=10$ event types in total.  We work on a subset of $2000$ sequences of taxi pickup events with average length $39$ and then end up with $K=10$ event types. We randomly sampled disjoint train, dev and test sets with $1400$, $200$ and $400$ sequences from the dataset. Given the average inter-arrival time $0.22$ (time unit is $1$ hour), we choose the prediction horizon as $4.5$ that approximately has $20$ event tokens per sequence.

{\bfseries StackOverflow } \citep{snapnets}. This dataset has two years of user awards on a question-answering website: each user received a sequence of badges and there are $K=22$ different kinds of badges in total. 
We randomly sampled disjoint train, dev and test sets with $1400,400$ and $400$ sequences from the dataset.
The time unit is $11$ days; the average inter-arrival time is $0.95$ and we set the prediction horizon to be $20$ that approximately covers $20$ event tokens.

\Cref{tab:stats_dataset} shows statistics about each dataset mentioned above.

\begin{table*}[tb]
	\begin{center}
		\begin{small}
			\begin{sc}
				\begin{tabularx}{1.00\textwidth}{l *{1}{S}*{3}{R}*{3}{S}}
					\toprule
					Dataset & \multicolumn{1}{r}{$K$} & \multicolumn{3}{c}{\# of Event Tokens} & \multicolumn{3}{c}{Sequence Length} \\
					\cmidrule(lr){3-8}
					&  & Train & Dev & Test & Min & Mean & Max \\
					\midrule
					Taobao & $17$ & $75000$ & $12000$ & $30000$ & $58$ & $59$ & $59$ \\
					Taxi & $10$ & $56000$ & $10000$ & $16000$ & $38$ & $39$ & $39$ \\
                    StackOverflow & $22$& $91000$& $26000$ & $27000$ & $41$ & $65$ & $101$\\
					\bottomrule
				\end{tabularx}
			\end{sc}
		\end{small}
	\end{center}
	\caption{Statistics of each dataset.}
	\label{tab:stats_dataset}
\end{table*}

\subsection{Implementation Details}\label{app:imp}

All models are implemented using the PyTorch framework \citep{paszke-17-pytorch}.

For the implementation of NHP, AttNHP, and thinning algorithm, we used the code from  the  public  Github  repository at {\small \url{https://github.com/yangalan123/anhp-andtt}} \citep{yang-2022-transformer} with MIT License.

For DualTPP, we used the code from the public Github repository at {\small \url{https://github.com/pratham16cse/DualTPP}} \citep{Deshpande_2021} with no license specified.

For the optimal transport distance, we used the code from the public Github repository at {\small \url{https://github.com/hongyuanmei/neural-hawkes-particle-smoothing}} \citep{mei-19-smoothing} with BSD 3-Clause License.

Our code can be found at {\small \url{https://github.com/alipay/hypro_tpp}} and {\small \url{https://github.com/iLampard/hypro_tpp}}.

\subsection{Training and Testing Details}\label{app:training_details}

{\bfseries Training Generators.} For AttNHP, the main hyperparameters to tune are the hidden dimension $D$ of the neural network and the number of layers $L$ of the attention structure. In practice, the optimal $D$ for a model was usually $32$ or $64$; the optimal $L$ was usually $1,2,3,4$. In the experiment, we set $D=32, L=2$ for AttNHP and $D=32, L=4$ for AttNHP-LG. To train the parameters for a given generator, we performed early stopping based on log-likelihood on the held-out dev set. 

%We set the maximum training epoch number of LG-CTTX twice as that of BS-CTTX.

{\bfseries Training Energy Function.} The energy function is built on NHP or AttNHP with $3$ MLP layers to project the hidden states into a scalar energy value. AttNHP is set to have the same structure as the base generator 'Att'. NHP is set to have $D=36$ so that the joint model have the comparable number of parameters with other competitors. During training, each pair of training sample contains $1$ positive sample and $5$ negative samples ($N=5$ in equation \ref{eqn:bin_entropy_loss} and \ref{eqn:multi_entropy_loss}), generated from generators. Regarding the regularization term in equation \ref{eqn:reg}, we choose $\beta=1.0$.

All models are optimized using Adam \citep{kingma-15}.

{\bfseries Testing.} During testing, for efficiency, we generates $20$ samples ($M=20$ in \cref{alg:joint_sampling}) per test prefix and select the one with the highest weight as the prediction. Increasing $M$ could possibly improves the prediction performance. 

{\bfseries Computation Cost.} 
All the experiments were conducted on a server with $256$G RAM, a $64$ logical cores CPU (Intel(R) Xeon(R) Platinum 8163 CPU @ 2.50GHz) and one NVIDIA Tesla P100 GPU for acceleration. On all the datasets, the training time of HYPRO-A and HYPRO-N is $0.005$ seconds per positive sequence.

For training, our batch size is 32. For Taobao and Taxi dataset, training the baseline NHP, NHP-lg, AttNHP, AttNHP-lg approximately takes 1 hour, 1.3 hour, 2 hours, and 3 hours, respectively (12, 16, 25, 38 milliseconds per sequence), training the continuous-time LSTM energy function and continuous-time Transformer energy function takes 20 minutes and 35 minutes (4 and 7 milliseconds per sequence pair) respectively.

For inference, inference with energy functions takes roughly 2 to 4 milliseconds. It takes ~0.2 seconds to draw a sequence from the autoregressive base model. 
Our implementation can draw multiple sequences at a time in parallel: it takes only about 0.4 seconds to draw $20$ sequences---only twice as drawing a single sequence. 
We have released this implementation. 

\begin{table*}[tb]
	\begin{center}
		\begin{small}
			\begin{sc}
				\begin{tabularx}{1.00\textwidth}{l *{2}{S}*{3}{S}}
					\toprule
					Model & \multicolumn{2}{c}{Description} & \multicolumn{3}{c}{Value used}\\
					\cmidrule(lr){4-6}
					& &  & Taobao & Taxi & StackOverflow \\
					\midrule
					DualTPP& \multicolumn{2}{c}{RNN hidden size} & $76$ & $76$ & $76$\\
					       & \multicolumn{2}{c}{Temporal embedding size} & $32$ & $32$ & $32$\\
					\hline
					NHP & \multicolumn{2}{c}{RNN hidden size} & $36$ & $36$ & $36$\\
					 \hline 
					 NHP-LG & \multicolumn{2}{c}{RNN hidden size} & $52$ & $52$& $52$\\
					 \hline 
					AttNHP & \multicolumn{2}{c}{Temporal embedding size} & $64$ & $64$& $64$\\
					 & \multicolumn{2}{c}{Encoder/decoder hidden size} & $32$ & $32$& $32$\\
					 & \multicolumn{2}{c}{Layers number} & $2$ & $2$& $2$\\

					 \hline
					AttNHP-LG & \multicolumn{2}{c}{Temporal embedding size} & $64$ & $64$ & $64$\\
					 & \multicolumn{2}{c}{Encoder/decoder hidden size} & $32$ & $32$ & $32$\\
					 & \multicolumn{2}{c}{Layers number} & $4$ & $4$ & $4$\\
				% 	 & \multicolumn{2}{c}{Training epoch} & $200$ & $200$\\
					 \hline 
					 HYPRO-N-B & \multicolumn{2}{c}{RNN hidden size in NHP} & $32$ & $32$ & $32$\\
					 \hline 
					 HYPRO-N-M & \multicolumn{2}{c}{RNN hidden size in NHP} & $32$ & $32$ & $32$\\
					 \hline 
					 HYPRO-A-B&  \multicolumn{2}{c}{Energy function is a clone of AttNHP} & na & na & na\\
					 \hline 
					 HYPRO-A-M&  \multicolumn{2}{c}{Energy function is a clone of AttNHP} & na & na &na\\
					 \hline 
					\bottomrule
				\end{tabularx}
			\end{sc}
		\end{small}
	\end{center}
	\caption{Descriptions and values of hyperparameters used for models trained on the two datasets.}
	\label{tab:model_params}
\end{table*}

\begin{table*}[tb]
	\begin{center}
		\begin{small}
			\begin{sc}
				 \begin{tabularx}{0.8\textwidth}{l*{3}{S}}
					\toprule
					Model  & \multicolumn{3}{c}{$\#$ of Parameters}\\
					\cmidrule(lr){2-4}
					& Taobao & Taxi & StackOverflow \\
					\midrule
					DualTPP  & $40.0$k & $40.1$k &$40.3$k \\
					NHP  & $19.6$k & $19.3$k &  $20.0$k\\
					NHP-LG  & $40.0$k & $39.3$k & $40.6$k \\
					AttNHP  & $19.7$k & $19.3$k & $20.1$k\\
					AttNHP-LG  & $38.3$k & $37.9$k & $38.7$k\\
					HYPRO-A-B  & $40.0$k & $40.5$k & $41.0$k\\
					HYPRO-A-M  & $40.0$k & $40.5$k &  $41.0$k \\
					 \hline 
					\bottomrule
				\end{tabularx}
			\end{sc}
		\end{small}
	\end{center}
	\caption{Total number of parameters for models trained on the three datasets.}
	\label{tab:model_size}
\end{table*}

\subsection{More OTD Results} \label{app:exp_results}
The optimal transport distance (OTD) depends on the hyperparameter $C_{\text{del}}$, which is the cost of deleting or adding an event token of any type. 
In our experiments, we used a range of values of $C_{\text{del}} \in \{0.05, 0.5, 1, 1.5, 2, 3, 4\}$, and report the averaged OTD in \cref{fig:main_results}. 

\begin{figure}[h]
\centering
        \begin{subfigure}[t]{0.99\linewidth}
	
            \includegraphics[width=1.0\linewidth]{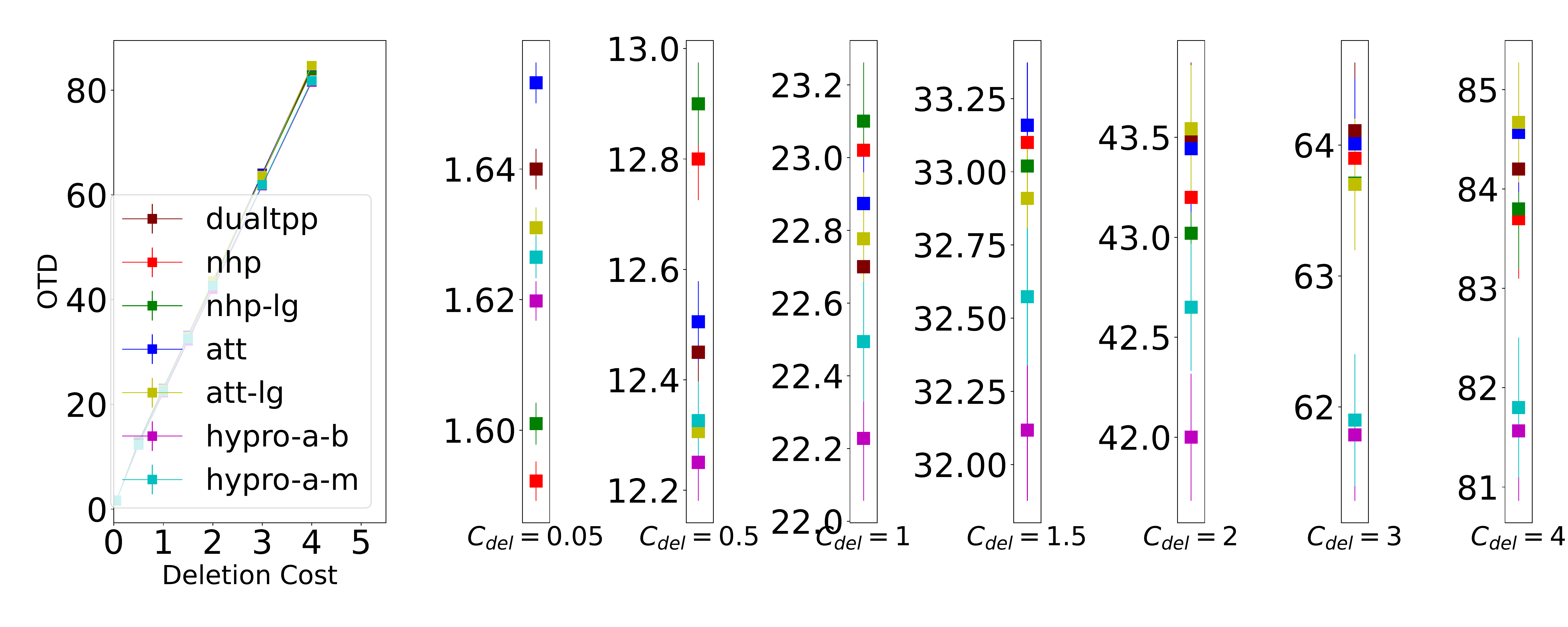}
			\vspace{-20pt}
			\caption{Taobao Data}\label{fig:results_taobao_otd_details}
		\end{subfigure}
		
		\begin{subfigure}[t]{0.99\linewidth}
		
			\includegraphics[width=1.0\linewidth]{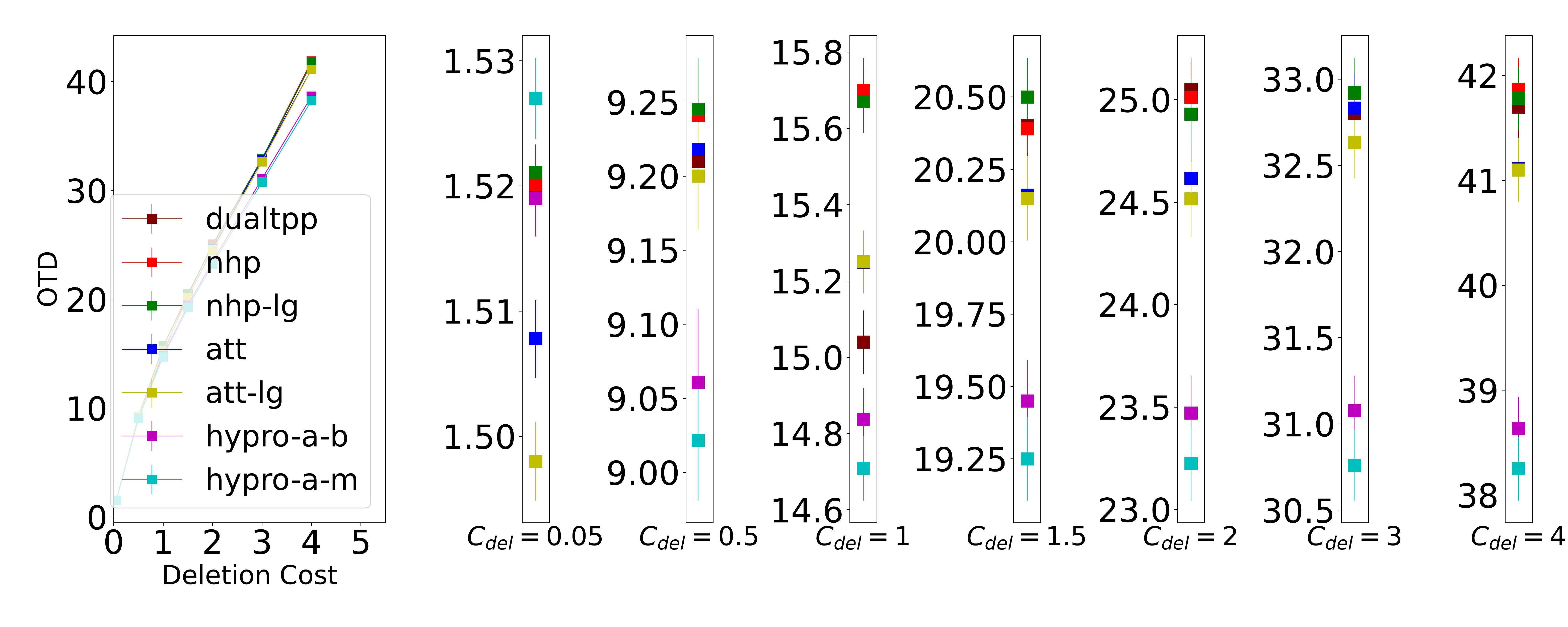}

			\vspace{-20pt}
			\caption{Taxi Data}\label{fig:results_taxi_otd_details}
		\end{subfigure}
            \begin{subfigure}[t]{0.99\linewidth}
			
			\includegraphics[width=1.0\linewidth]{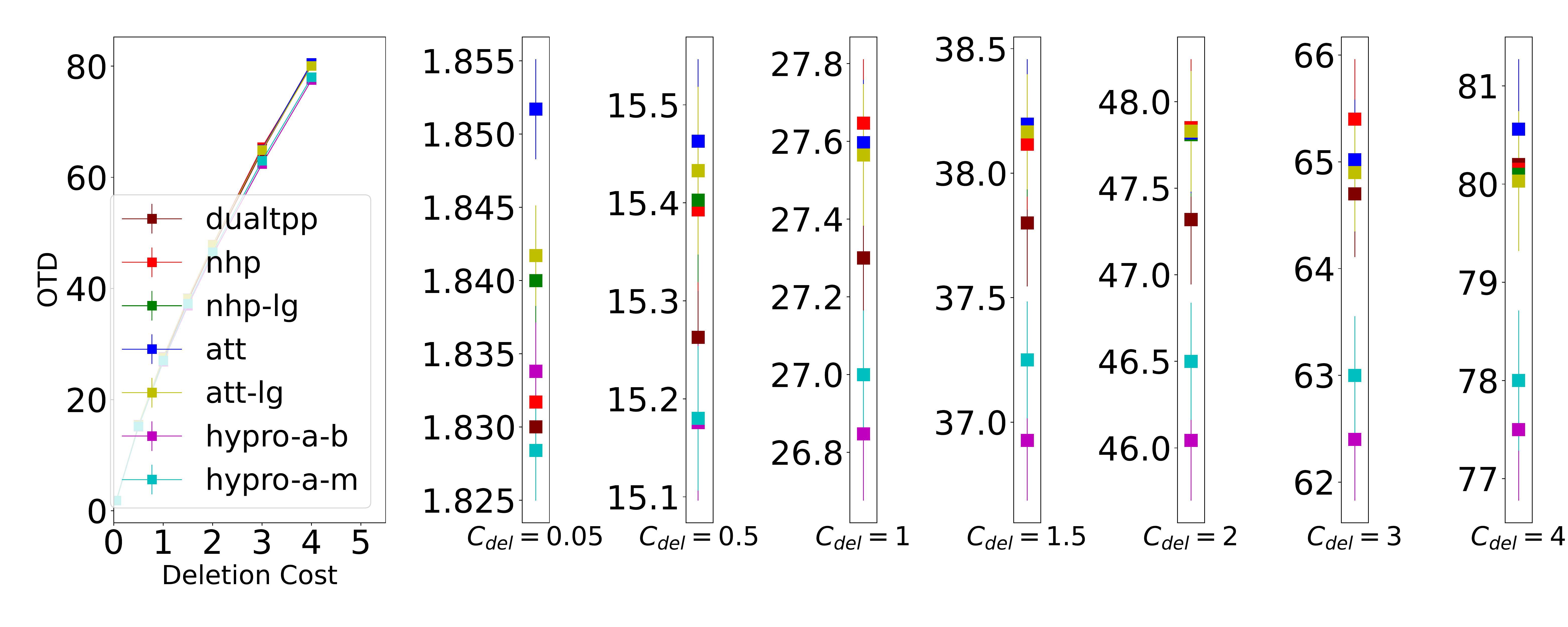}

			\vspace{-20pt}
			\caption{StackOverflow Data}\label{fig:results_so_otd_details}
		\end{subfigure}
		\caption{OTD for each specific deletion/addition cost $C_{\text{del}}$.}
		\label{fig:results_otd_details}
\end{figure}

In this section, we show the OTD for each specific $C_{\text{del}}$ in \cref{fig:results_otd_details}. 
As we can see, for all the values of $C_{\text{del}}$, our HYPRO method consistently outperforms the other methods.

\subsection{Analysis Details: Baseline That Ranks Sequences by the Base Model}\label{app:newbaseline}
To further verify the usefulness of the energy function in our model, we developed an extra baseline method that ranks the completed sequences based on their probabilities under the base model, from which the continuations were drawn. This baseline is similar to our proposed HYPRO framework but its scorer is the base model itself.

\begin{figure*}[!htb]
	\begin{center}
		\includegraphics[width=0.47\linewidth]{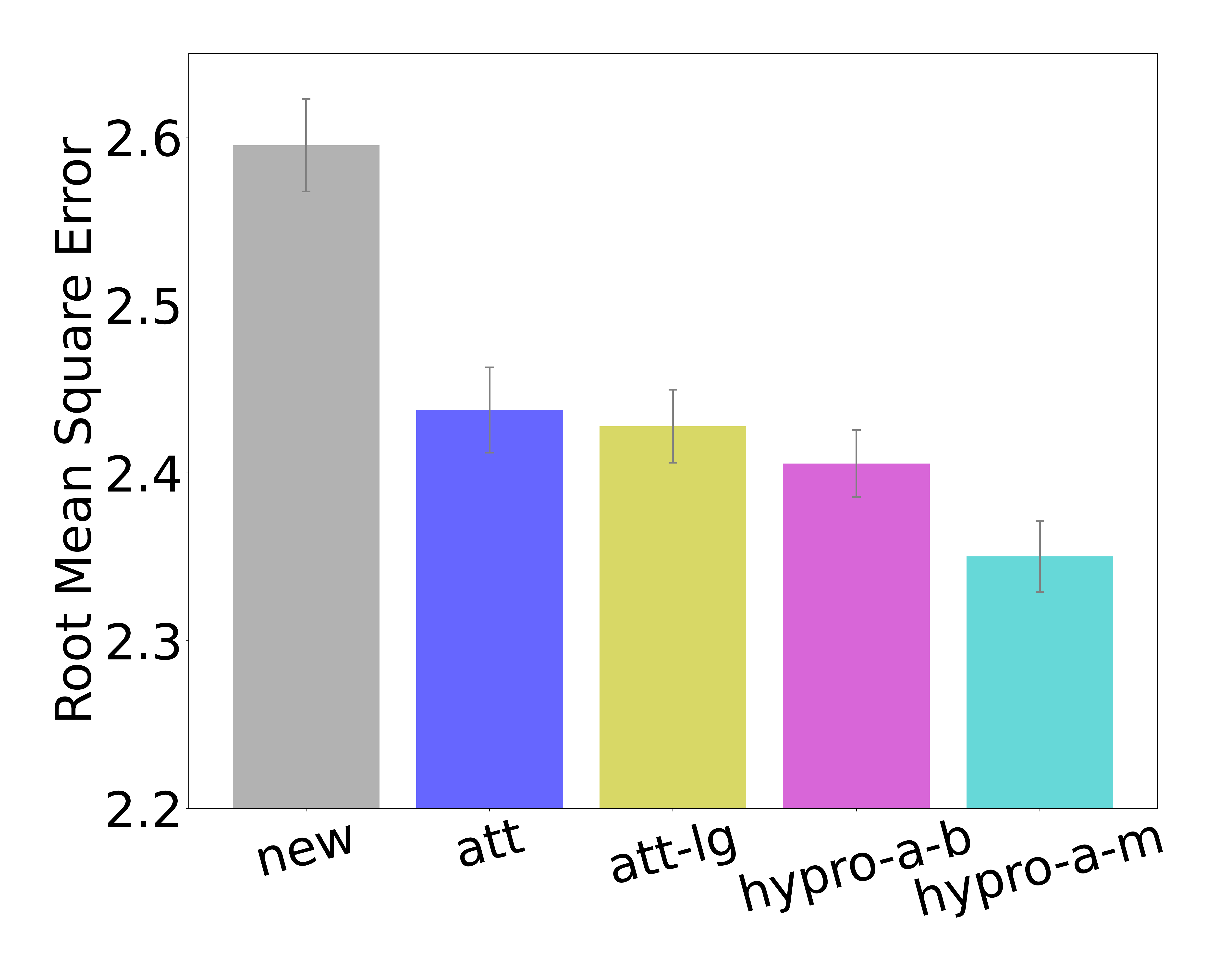}
		~
		\includegraphics[width=0.47\linewidth]{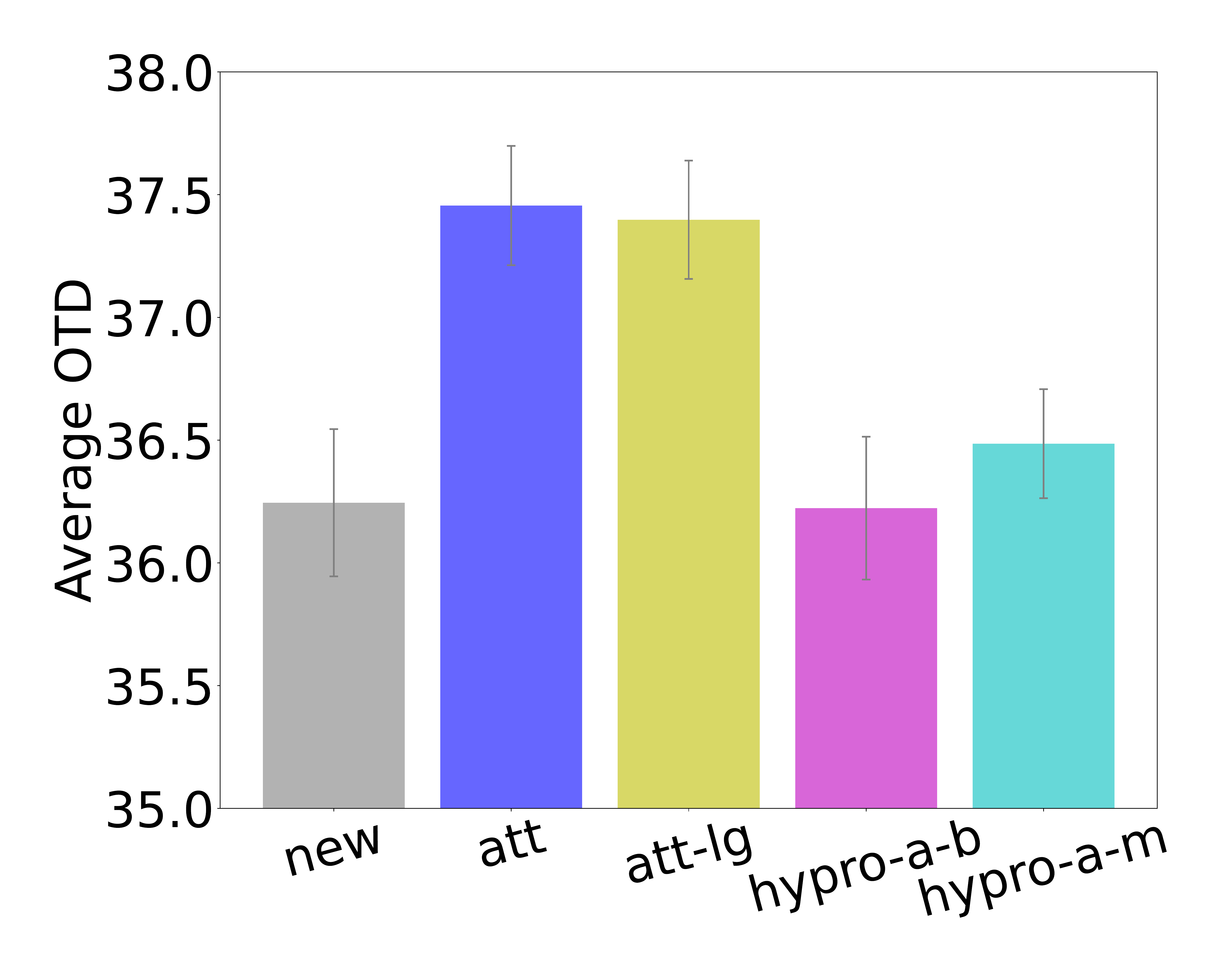}
		\vspace{-10pt}
		\caption{Evaluation of new baseline on Taobao dataset. The base model is AttNHP. The performances of the other methods are copied from \cref{fig:main_taobao}.}\label{fig:new_baseline}
	\end{center}

\end{figure*}

We evaluated this baseline on the Taobao dataset. The results are in \cref{fig:new_baseline}. 
As we can see, this new baseline method is no better than our method in terms of the OTD metric but much worse than all the other methods in terms of the RMSE metric. 

\subsection{Analysis Details: Statistical Significance}\label{app:sigtest}
We performed the paired permutation test to validate the significance of our proposed regularization technique. Particularly, for each model variant (hypro-a-b or hypro-a-m), we split the test data into ten folds and collected the paired test results with and without the regularization technique for each fold. Then we performed the test and computed the p-value following the recipe at \url{https://axon.cs.byu.edu/Dan/478/assignments/permutation_test.php}. 

\begin{figure*}[!htb]
	\begin{center}
	    \begin{subfigure}[t]{0.49\linewidth}
			   \includegraphics[width=0.47\linewidth]{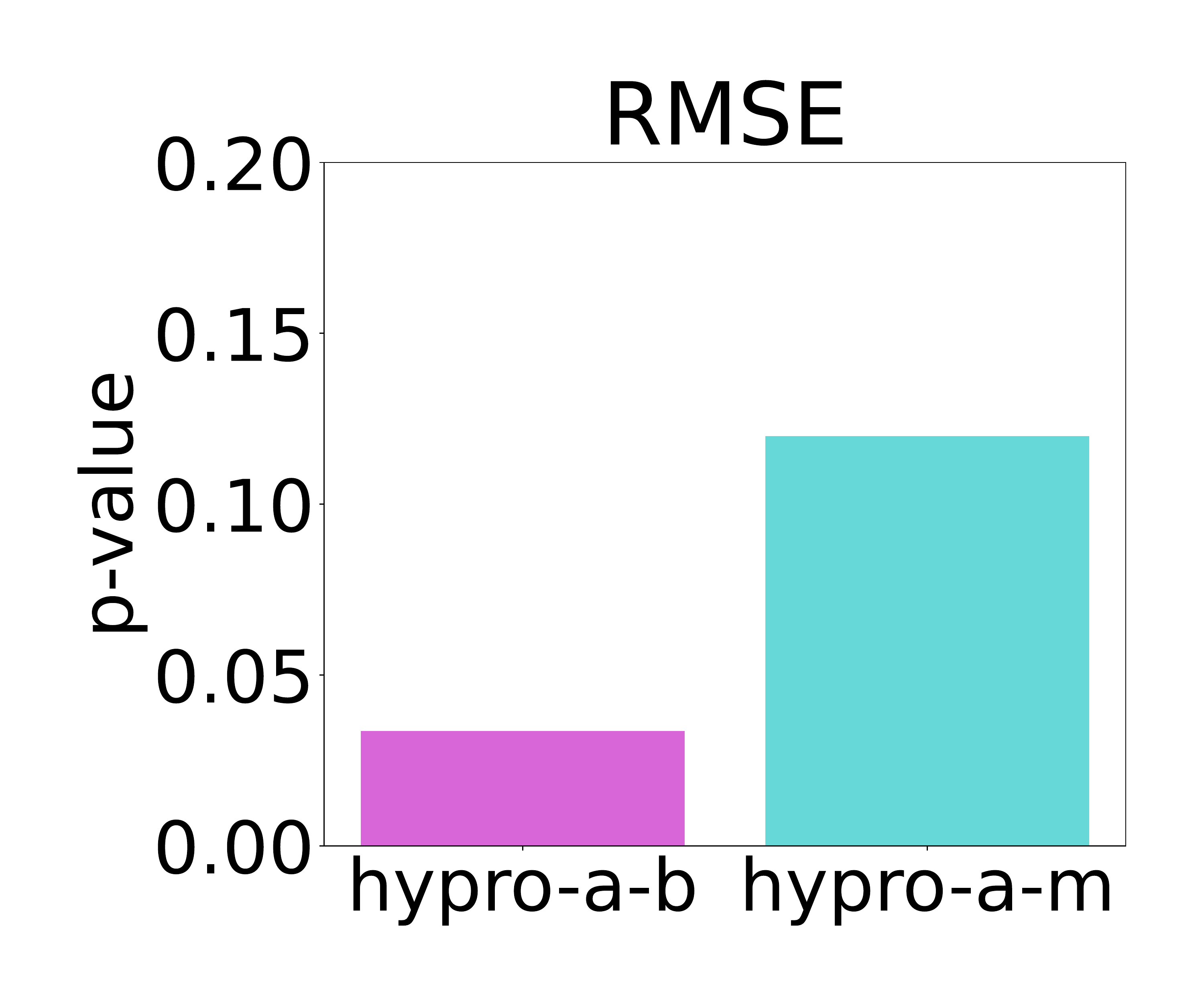}
			~
			\includegraphics[width=0.47\linewidth]{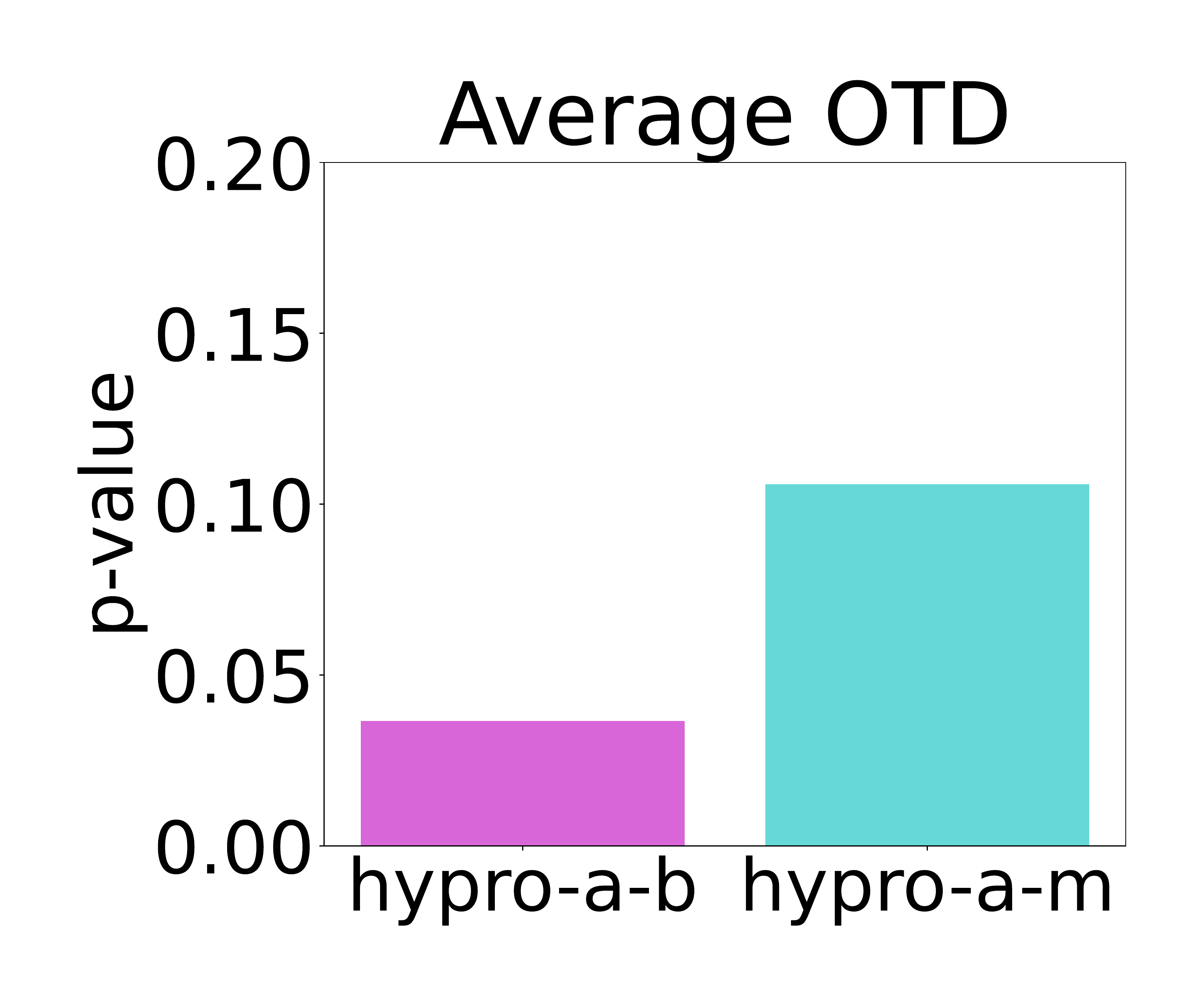}
			\vspace{-10pt}
			\caption{Taobao dataset.}\label{fig:perm_taobao}
		\end{subfigure}
		~
		\begin{subfigure}[t]{0.49\linewidth}
			
			\includegraphics[width=0.47\linewidth]{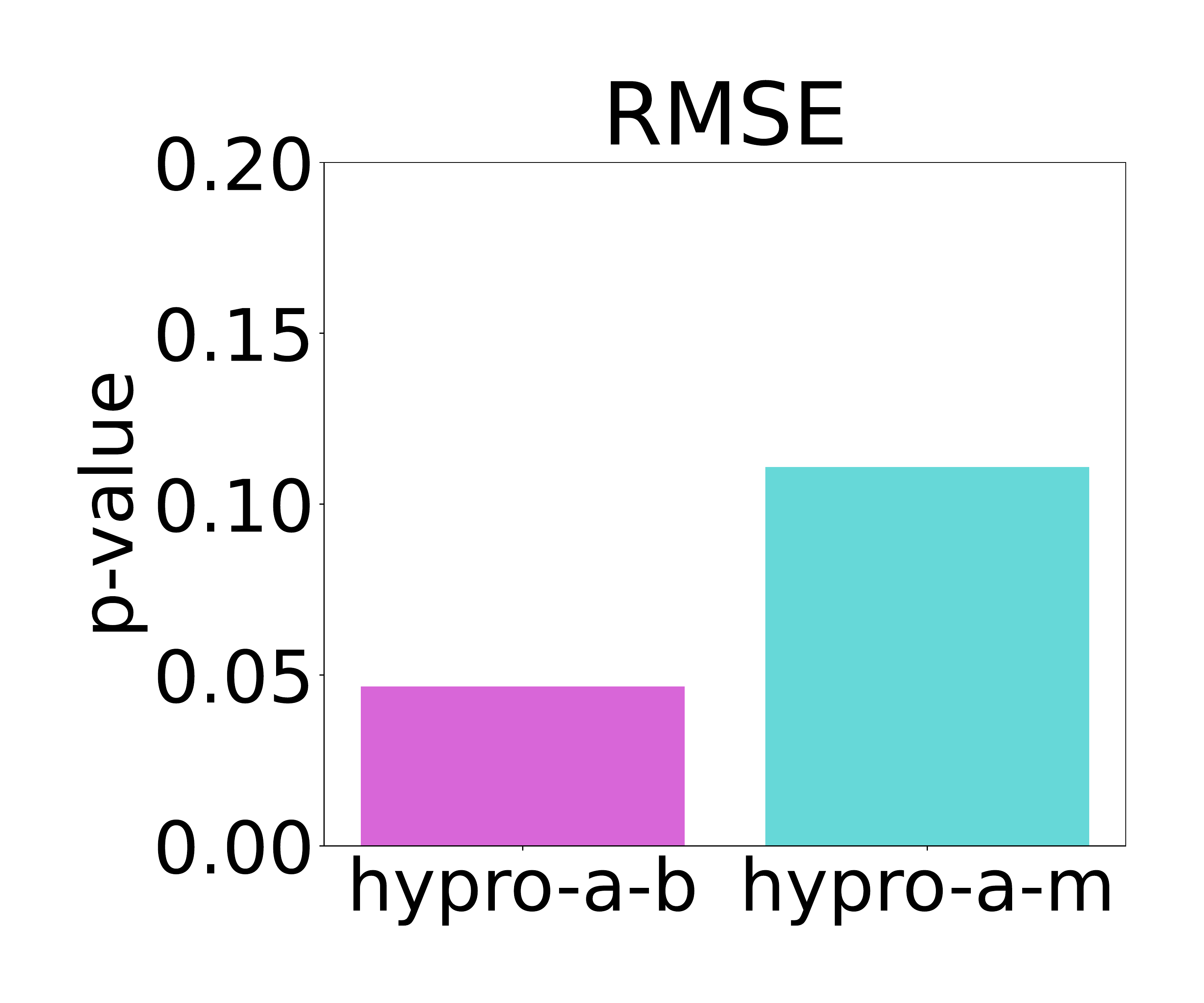}
			~
			\includegraphics[width=0.47\linewidth]{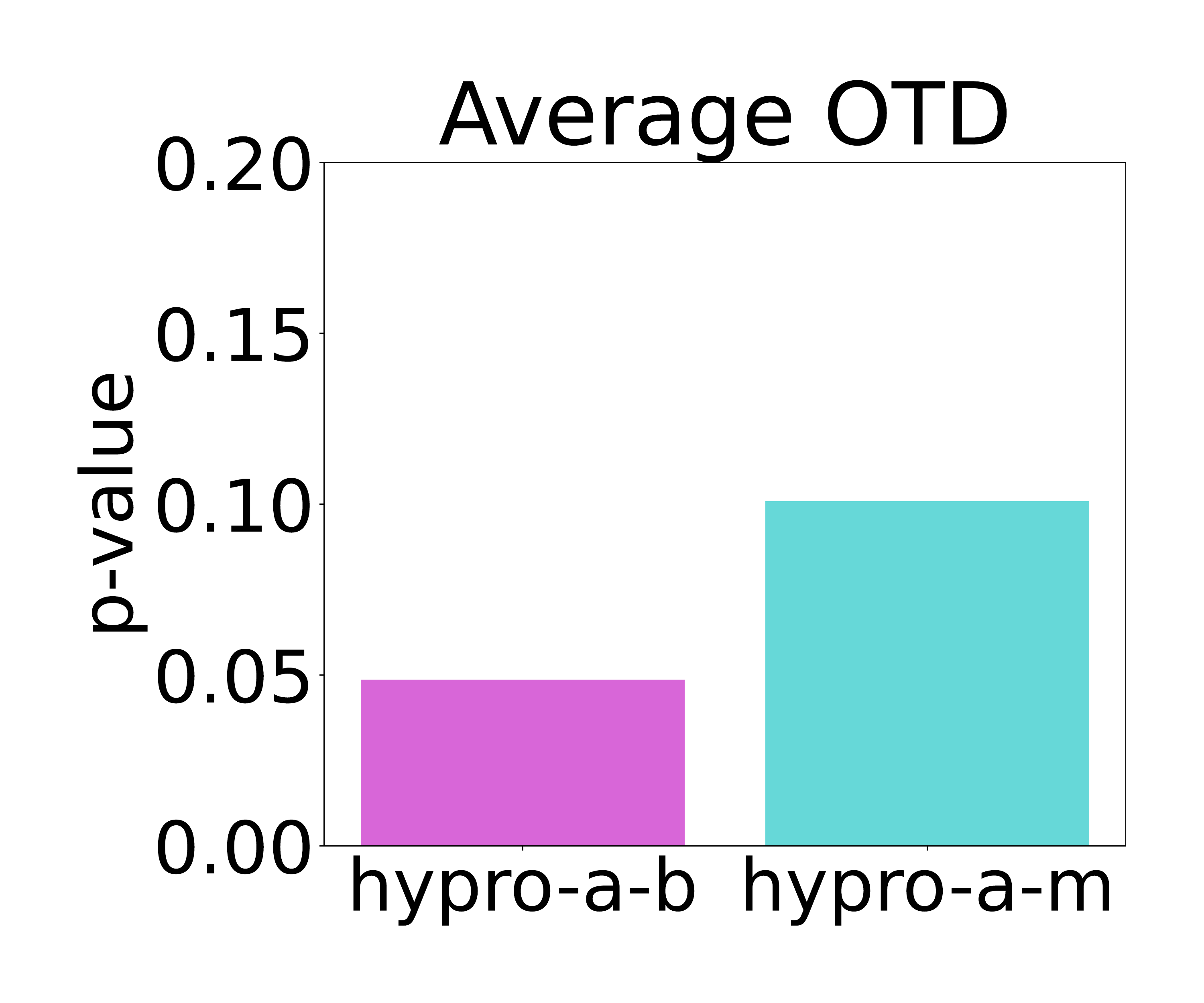}
			\vspace{-10pt}
			\caption{Taxi dataset.}\label{fig:perm_taxi}
		\end{subfigure}
		\vspace{-4pt}
		\caption{Statistical significance of our regularization on the Taobao and Taxi datasets.}\label{fig:perm}
	\end{center}
\end{figure*}

The results are in \cref{fig:perm}. It turns out that the performance differences are strongly significant for hypro-a-b (p-value $<0.05$ ) but not significant for hypro-a-m (p-value $\approx 0.1$ ). This is consistent with the findings in \cref{fig:energy_distribution}. 

\end{document}